\documentclass[journal]{IEEEtran}
\usepackage{amsmath,amsfonts}
\usepackage{algorithmic}
\usepackage{algorithm}
\usepackage{array}
\usepackage[caption=false,font=normalsize,labelfont=sf,textfont=sf]{subfig}
\usepackage{textcomp}
\usepackage{stfloats}
\usepackage{url}
\usepackage{verbatim}
\usepackage{graphicx}
\usepackage{cite}
\usepackage[numbers,sort&compress]{natbib}
\usepackage{booktabs} 
\usepackage{xcolor}
\begin{document}
%A Dual-branch Interactive Semantic Segmentation Network For Relic Landslide Detection
\title{MRIFE: A Mask-Recovering and Interactive-Feature-Enhancing Semantic Segmentation Network For Relic Landslide Detection
}

\author{Juefei He, Yuexing Peng,~\IEEEmembership{Member,~IEEE,} Wei Li,~\IEEEmembership{Senior Member,~IEEE,} Junchuan Yu, Daqing Ge, 

and Wei Xiang,~\IEEEmembership{Senior Member,~IEEE}
        % <-this % stops a space
\thanks{Juefei He and Yuexing Peng are with the key lab of Universal Wireless Communication,
MoE, School of Information and Communication Engineering, Beijing University of Posts and Telecommunications, Beijing 100876, China (e-mail: hejuefei2018@bupt.edu.cn; yxpeng@bupt.edu.cn).}% <-this % stops a space
\thanks{Wei Li is with the School of Information and Electronics, Beijing Institute
of Technology, Beijing 100081, China (e-mail: liwei089@ieee.org).}
\thanks{Junchuan Yu and Daqing Ge are with the Department of Satellite Application Research, China Aero Geophysical Survey and Remote Sensing Center for Natural Resources, Beijing 100083, China (e-mail: yujunchuan@mail.cgs.gov.cn; gedaqing@mail.cgs.gov.cn).}
\thanks{Wei Xiang is with the School of Computing, Engineering and Mathematical Sciences, La Trobe University, Melbourne, VIC 3086, Australia (e-mail: w.xiang@latrobe.edu.au).}}

\markboth{Journal of \LaTeX\ Class Files,~Vol.~14, No.~8, August~2021}%
{Shell \MakeLowercase{\textit{et al.}}: A Sample Article Using IEEEtran.cls for IEEE Journals}


\maketitle

\begin{abstract}
Relic landslide, formed over a long period, possess the \textcolor{blue}{potential} for reactivation, making them a hazardous geological phenomenon. While reliable relic landslide detection \textcolor{blue}{benefits} the effective monitoring and prevention of landslide disaster, \textcolor{blue}{semantic segmentation using} high-resolution remote sensing \textcolor{blue}{images} for relic landslides \textcolor{blue}{faces} many challenges, including \textcolor{blue}{the} object visual blur problem, due to the changes of appearance caused by prolonged natural evolution and human activities, and \textcolor{blue}{the} small-sized dataset problem, due to difficulty in recognizing and labelling the samples. To address these challenges, a semantic segmentation model, termed mask-recovering and interactive-feature-enhancing (MRIFE), is proposed for more efficient feature extraction and separation. \textcolor{blue}{Specifically, a contrastive learning and mask reconstruction method with locally significant feature enhancement is proposed to improve the ability to distinguish between the target and background and represent landslide semantic features. Meanwhile, a dual-branch interactive feature enhancement architecture is used to enrich the extracted features and address the issue of visual ambiguity. Self-distillation learning is introduced to leverage the feature diversity both within and between samples for contrastive learning, improving sample utilization, accelerating model convergence, and effectively addressing the problem of the small-sized dataset.} The proposed MRIFE is evaluated on a real relic landslide dataset, and experimental results show that it greatly improves the performance of relic landslide detection. For the semantic segmentation task, compared to the baseline, the precision increases from 0.4226 to 0.5347, the mean intersection over union (IoU) increases from 0.6405 to 0.6680, the landslide IoU increases from 0.3381 to 0.3934, and the F1-score increases from 0.5054 to 0.5646.

\end{abstract}


\section{Introduction}
\IEEEPARstart
{R}{elic landslides} is the result of prolonged and intricate geological processes occurring on slopes \cite{1}. Although the majority of relic landslides exhibit long-term stability, triggers such as human activities, earthquakes, and rainfall can lead to the reactivation and renewed sliding of these relic landslides. Relic \textcolor{blue}{landslides} has resulted in significant harm to both human life and property safety, as well as the natural environment \cite{2,3}. From the 1950s to the 1970s, more than 170 large and medium-sized landslides occurred along the 98 km from Baoji Gorge to Changxing, nearly half of which were relic landslides \cite{4}. In order to reduce the losses caused by the reactivation of relic landslides, it is imperative to undertake extensive detection and monitoring of relic landslides on a large scale.

Traditional landslide detection methods rely on expert interpretation and field surveys. Detection results are obtained by analyzing the geomorphic features of the geological disaster area, as well as the mechanisms and processes of landslide occurrence \cite{5}. However, the manual interpretation process is complex, time-consuming, and labor-intensive, relying \textcolor{blue}{heavily} on expert experience. Meanwhile, landslides have diverse causes and variable manifestations, leading to inconsistencies in detection results and significant fluctuations in accuracy. Moreover, it is difficult to meet the requirements \textcolor{blue}{for} wide-area automatic and reliable detection \cite{6,7}.

With the advancement of high-resolution remote sensing technology, a wealth of ground observation data has emerged, including high-resolution remote sensing images (HRSIs), digital elevation models (DEMs), and interferometric synthetic aperture radar (InSAR) data \cite{8,9,10}.  Detecting landslides \textcolor{blue}{using} remote sensing technology has become a trend.

Classic machine learning methods for classification and information extraction from HRSIs include two categories: pixel-level and object-level \cite{11}. The former classify each pixel directly \cite{12,13,14,15,16,17}, while the latter use Object-Based Image Analysis to classify objects with similar spectral, spatial, and hierarchical features for landslide detection \cite{18,19,20,21,22}. Although machine learning methods save time and labor, their accuracy \textcolor{blue}{heavily} relies on the selection of sample features and tuning of hyperparameters, such as segmentation thresholds and scales, which results in poor generalization performance and severely limits their wide-area application.

Deep learning methods, with their superior abstraction and end-to-end learning capabilities, have significantly improved efficiency and accuracy \textcolor{blue}{of} landslide detection, such as R-CNN \cite{23,24,25}, FCN \cite{26}, DeepLab series \cite{27}, and other CNN-based models \cite{29,30,33}. The transformer is celebrated in natural language processing for its exceptional capability to capture global relationships, offering a promising approach to semantic segmentation. Recently, many \textcolor{blue}{studies} have proved that Transformers can achieve remarkable performance in computer vision \cite{35,36}. For large-scale automatic landslide detection, numerous advanced semantic segmentation networks have been proposed \cite{37,38,39}.

However, existing studies primarily \textcolor{blue}{focus} on new landslides, which exhibit clear color and texture differences from backgrounds, achieving high detection accuracy. Studies on relic landslides are much fewer, with much worse performance \cite{40,41}. There are still two great challenges for relic landslide detection using HRSIs, including: 1) Visual blur problem. Relic landslides, formed long time ago, have undergone changes due to prolonged natural environmental evolution and human activities, resulting in \textcolor{blue}{surfaces} that closely resemble those of the surrounding environment. \textcolor{blue}{Consequently,} in HRSI data, the optical features of landslides are blurred and the differences from non-landslide areas are very subtle. Both CNNs and Transformers struggle to capture such subtle differences; 2) Small-sized dataset problem. Constructing relic landslide dataset is extremely difficult due to the technical difficulty of accurate recognition of landslides and intensive time consumption on delineating the boundary of landslide. Small-sized dataset cannot sufficiently support powerful model due to overfitting, which imposes higher demands on the learning and generalization capabilities

In this paper, we propose a mask-recovering and interactive-feature-enhancing (MRIFE) semantic segmentation network to detect visually blurred relic landslides \textcolor{blue}{in} a small-sized HRSI dataset. \textcolor{blue}{The MRIFE model consists of two branches. In the feature enhancement branch, we perform targeted mask reconstruction on key edge areas, such as the landslide sidewalls and rear walls, at the feature map level to enhance the model's feature representation capabilities. Supervised contrastive learning is applied to distinguish between similar features of the landslide edge and the background. Positive samples are derived from the landslide boundary, while negative samples are extracted from the environmental background, allowing the model to focus on the subtle differences between similar features. Additionally, a self-distillation framework is employed to facilitate both within-sample reconstruction learning and cross-sample contrastive learning. This approach leverages the feature diversity within and between samples, enabling the model to learn broader patterns and more nuanced features. By complementing the primary features extracted by the segmentation branch with the fine-grained features from the feature enhancement branch, the MRIFE model significantly improves pixel-level classification accuracy.} 

The main contributions of this paper are as follows:
\begin{itemize}
\item {
We present a detector called MRIFE, which is designed to address the visual blur problem and small-sized dataset problem \textcolor{blue}{in} relic landslides detection using HRSIs.
}
\item {
A dual-branch interactive feature enhancement framework is proposed to achieve highly efficient feature fusion through multi-task training and mutual guidance between \textcolor{blue}{the} two branches, enriching the semantic information.
}
\item {
A masked feature modeling (MFM) method is introduced to \textcolor{blue}{learn} high-level features by masking and recovering key features of landslide targets via self-distillation learning, improving the efficiency of feature extraction.
}
\item {
To distinguish between landslides and background, a semantic feature contrast enhancement (SFCE) method is developed, which performs supervised contrastive learning between pixel blocks of landslide \textcolor{blue}{edges} and background to facilitate feature separation in semantic space.
}
\item {
We elaborately design experiments to \textcolor{blue}{demonstrate} the effectiveness of our proposed method. Compared to the baseline, our model achieved a 5.5\% performance improvement in the key metric landslide-IoU.
}
\end{itemize}

\section{Related Work}

\subsection{Masked Image Modeling}
In masked image modeling (MIM), parts of an image are masked (i.e., obscured), and the model attempts to reconstruct the obscured portions from the remaining visible parts. This approach helps the model learn the intrinsic structure and features of images. In \cite{42}, the SimMIM algorithm adopts a straightforward random masking strategy and utilizes an encoder-decoder structure to predict the pixel values of the masked regions. This approach is simple, effective, and achieves good performance. Our method is inspired by SimMIM but differs in that we selectively mask portions of landslide edges and background based on labels. Instead of using a decoder to reconstruct the image, we directly reconstruct the features corresponding to the masked positions on the feature map. \textcolor{blue}{This} way, our work learns more abstract and fundamental features, reducing the impact of visual blur problem.
\subsection{Contrastive Learning}
In \cite{43}, \textcolor{blue}{each} instance and its \textcolor{blue}{data-augmented versions} constitute the positive samples, while all the other samples constitute the negative ones. By comparing the similarities and differences of the contrastive samples, semantic features are extracted by a CNN and \textcolor{blue}{non-parametric} softmax. In our previous work \cite{44}, a sub-object-level contrastive learning is proposed, which learns features more effectively by using local information of the target to construct positive and negative samples. Unlike contrasting with object-level samples, we perform contrastive learning at the level of pixel blocks, utilizing landslide edge blocks to construct positive pairs and background blocks as negative samples to capture the subtle differences between target and non-target features more directly. Additionally, we \textcolor{blue}{incorporate} the reconstructed masked blocks into the sample pairs, further enhancing the efficiency of feature extraction.

\begin{figure*}[b]
\centering
\includegraphics[scale=0.4]{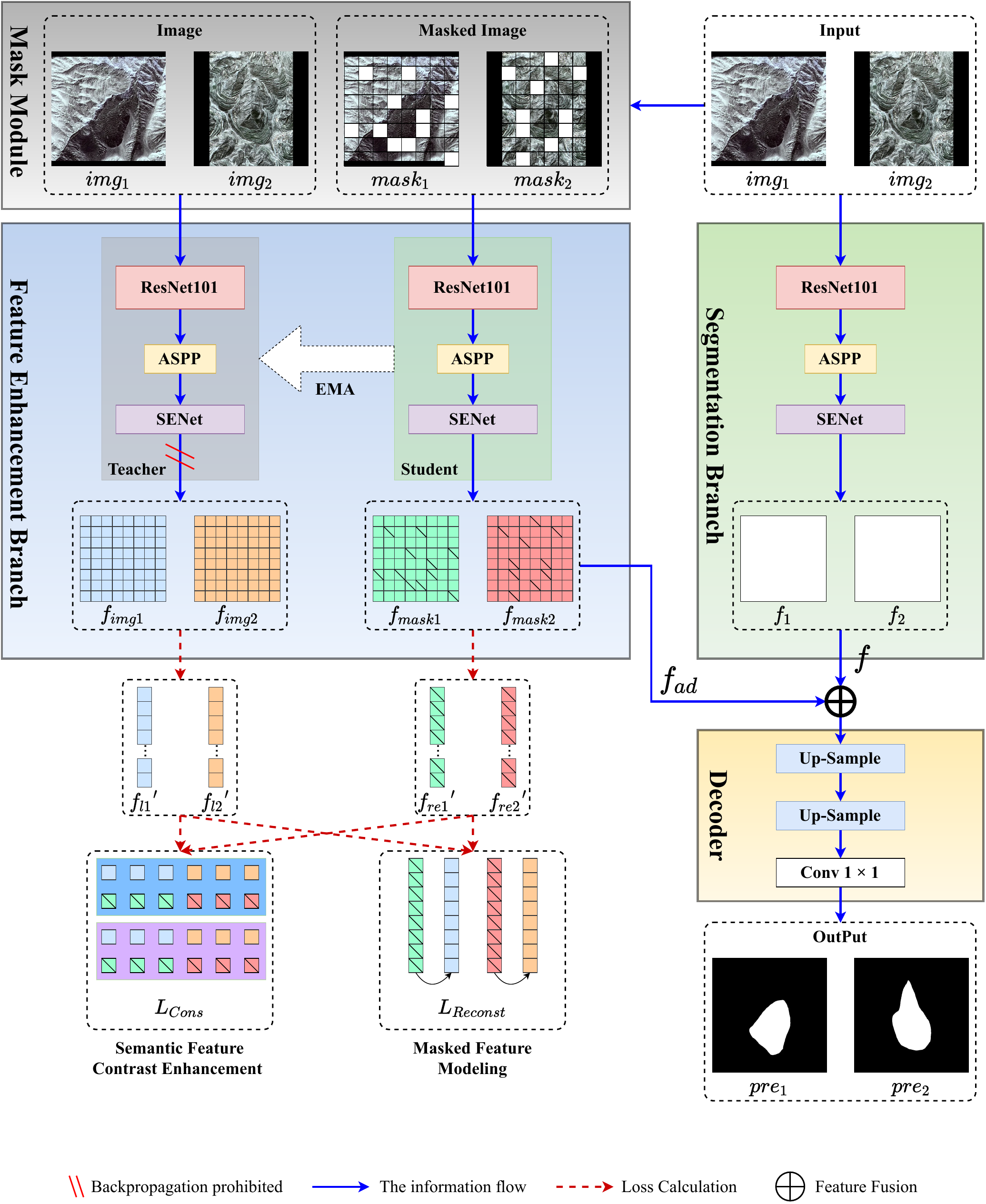}
\caption{Framework of the MRIFE. The two branches independently extract features and the decoder recovers the input images after feature fusion. A specially designed joint loss guides the updating of the two branches where the three encoders have the same network architecture but do not share parameters.}
\label{fig_1}
\end{figure*}
\subsection{Self-distillation}
The self-distillation method distills knowledge not from posterior distributions but \textcolor{blue}{from} past iterations of model itself and is cast as a discriminative self-supervised objective. \citet{45} propose a metric-learning formulation called BYOL, where features are trained by matching them to representations obtained with a momentum encoder. In \cite{46}, DINO is proposed, which leverages self-distillation through a teacher-student network to learn rich and robust feature representations from large amounts of unlabeled data, achieving efficient, stable, and superior performance in unsupervised tasks. Inspired by DINO, our approach utilizes self-distillation to perform the task of feature enhancement branch. The teacher network distills the average of the student networks, learning the commonalities among generations of \textcolor{blue}{students}, and provides guidance during student distillation, effectively suppressing overfitting issues caused by the small-sized dataset.
\subsection{Feature Fusion}
Attention mechanisms are mainstream for feature fusion. In \cite{47}, a self-attention mechanism \textcolor{blue}{is} introduced, which uses learnable queries (Q), keys (K), and values (V) to filter important information from features and discard less important details. In \cite{48}, channel attention is proposed, where the network autonomously learns to capture the significance of each channel in the feature maps. In \cite{49}, spatial attention is used to identify key spatial locations and enhance their feature representations. Our method for feature enhancement is more direct and efficient. Unlike approaches that learn weights, we use explicit task objectives to guide the independent feature extraction in both branches. By fusing features from the two branches, we achieve interactive and supplementary feature enhancement.

\section{Proposed Method}
We propose a dual-branch interactive feature-enhanced framework to realize relic landslide detection. In this section, we first specify the overall architecture, then provide detailed elaborations on each module of the network.

\subsection{Overview}
As illustrated in Fig. \ref{fig_1}, the proposed MRIFE model consists \textcolor{blue}{of} three functional modules, i.e., the segmentation branch, the feature enhancement branch, and the fusion module and decoder. Original and masked images are simultaneously input into the segmentation branch and the feature enhancement branch, where features are independently extracted and fused in the fusion module. Ultimately, the decoder predicts the segmentation results.

\subsection{Segmentation Branch}

Based on our previous work \cite{44}, the same feature extraction module is employed as the encoder, which consists of the backbone (ResNet101), ASPP (Atrous Spatial Pyramid Pooling), and SE \textcolor{blue}{(Squeeze-and-Excitation)} module for the extraction of high-dimensional semantic features from the input RGB image. The ResNet101 is employed as the backbone for feature extraction due to its proven effectiveness in feature extraction. It comprises four residual blocks, each of which has a different resolution of feature maps; The ASPP module is inserted for multi-scale feature fusion. By using the dilation convolution with different receptive fields to sample the feature maps, the encoder can effectively take into account both small-scale detailed texture information and large-scale overall morphology information simultaneously; SE module adaptively weights all channels in the feature map, allowing important channels to have greater weights and yielding a substantial enhancement in the capability to extract features.

\subsection{Feature Enhancement Branch}
The feature enhancement branch consists of a masking module and a teacher-student network. As shown in Fig. \ref{fig_1}, the masking module processes the inputs and passes them to the teacher-student network, where features are extracted \textcolor{blue}{by performing} masked feature modeling task and semantic feature contrast enhancement task.
\subsubsection{Mask Module}
\begin{figure}[ht]
\centering
\begin{tabular}{c@{\hspace{2pt}}c@{\hspace{2pt}}c@{\hspace{2pt}}c@{\hspace{2pt}}c}
%第1行
\includegraphics[width=0.08\textwidth]{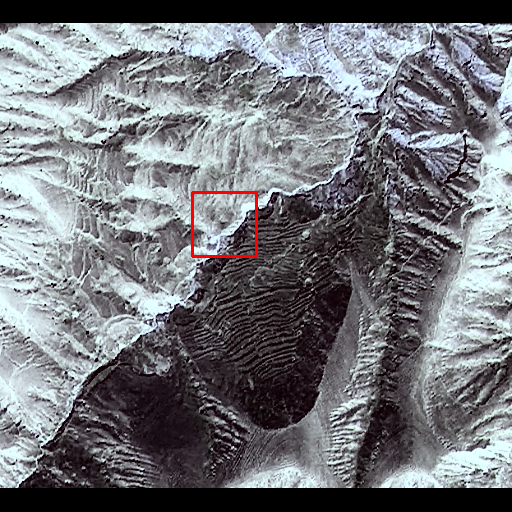} & 
\includegraphics[width=0.08\textwidth]{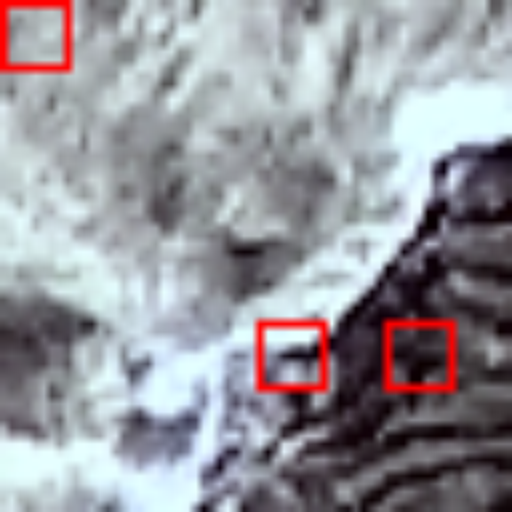} & 
\includegraphics[width=0.08\textwidth]{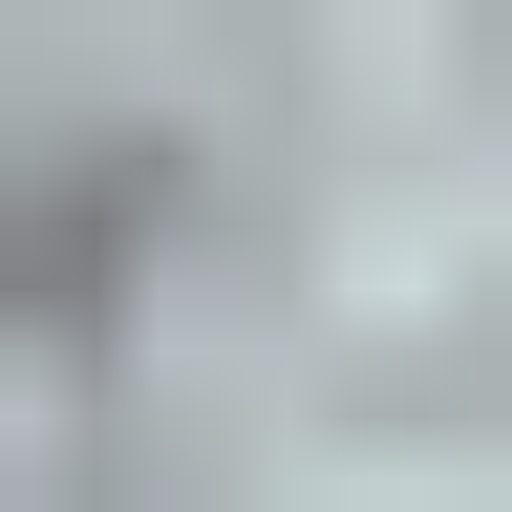} & 
\includegraphics[width=0.08\textwidth]{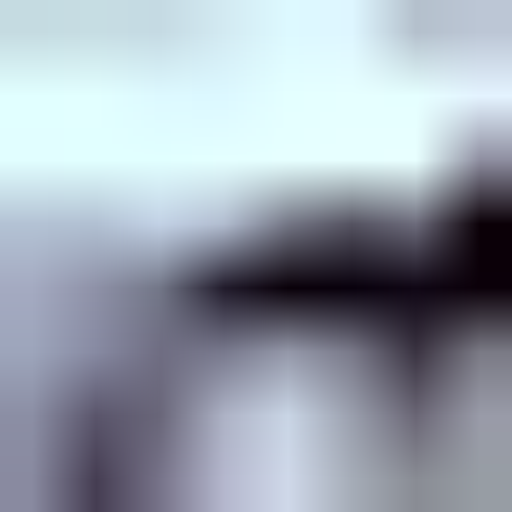} & 
\includegraphics[width=0.08\textwidth]{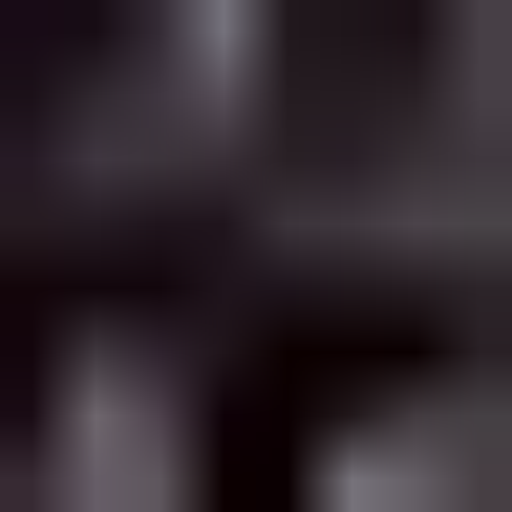} \\[-2pt]
%第2行
\includegraphics[width=0.08\textwidth]{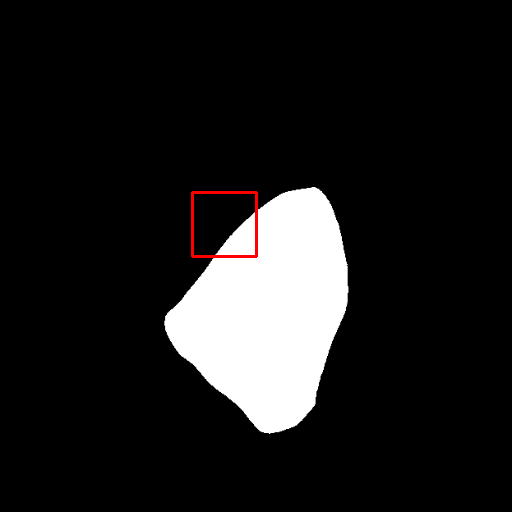} & 
\includegraphics[width=0.08\textwidth]{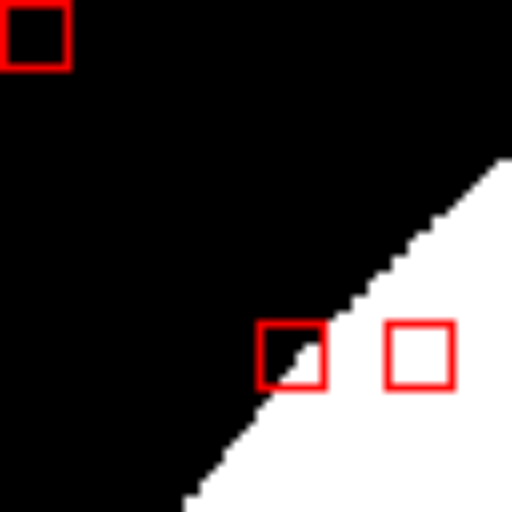} & 
\includegraphics[width=0.08\textwidth]{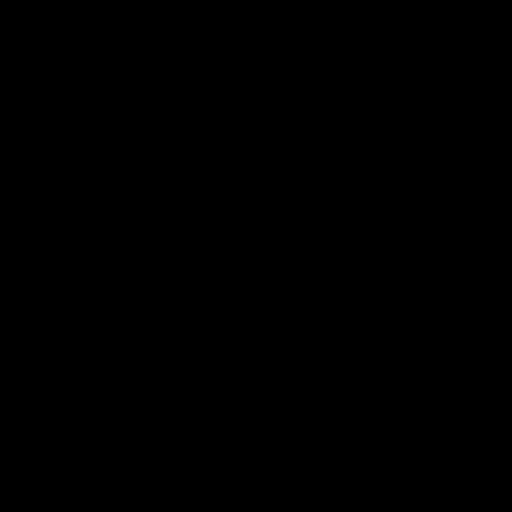} & 
\includegraphics[width=0.08\textwidth]{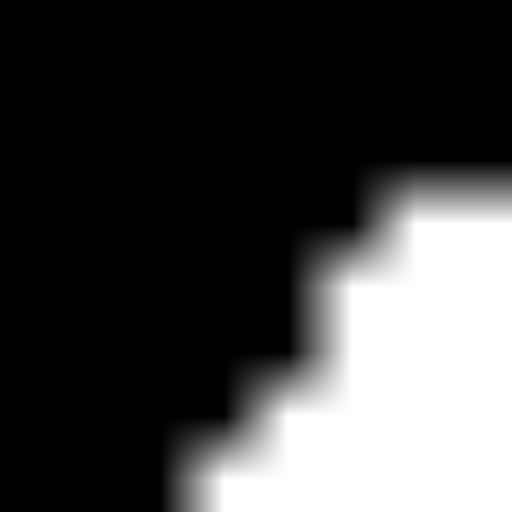} & 
\includegraphics[width=0.08\textwidth]{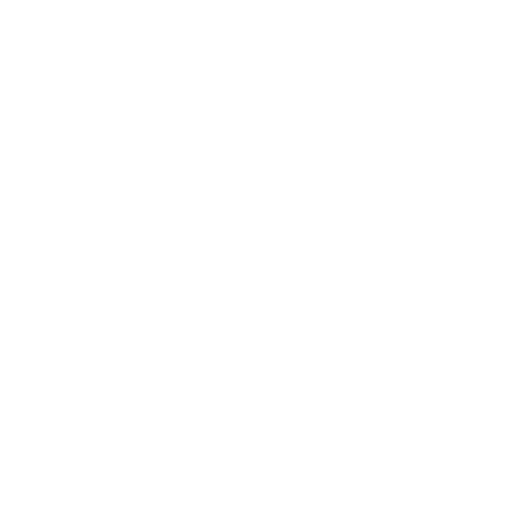} \\
(a) & (b) & (c) & (d) & (e) \\ 
\end{tabular}
\caption{(a) Original sample with $512\times512$ pixels; (b) part of original sample with $64\times64$ pixels; (c)-(e) masked block with $8\times8$ pixels: non-landslide block, landslide edge block, and landslide interior block.}
\label{fig_2}
\end{figure}

We partition the input image of size $512\times512$ into $64\times64$ blocks, each block consisting of $8\times8$ pixels. Since the downsampling rate of the feature extractor is $8$, each $8\times8$ block in the input image corresponds exactly to one feature point in the feature map. The mask module classifies the $8\times8$ blocks into three categories: non-landslide blocks, which contain fewer than $7$ landslide pixels; landslide edge blocks, which contain between $7$ and $57$ landslide pixels; and landslide interior blocks, which contain more than $57$ landslide pixels. It is worth noting that the landslide interior blocks exhibit features highly similar to the surrounding environment in the RGB image, contributing minimally to landslide detection. Consequently, the mask module directly discards them. This strategy is validated by the cross-validation experiment detailed in Section V.

For each image, the mask module randomly selects landslide edge blocks (i.e., class 1) and non-landslide blocks (i.e., class 0) for masking, in order to generate masked images that are paired with the original images. The masking process ensures an equal number of blocks between \textcolor{blue}{the} two classes. Additionally, the mask module records the positions of the masked landslide edge blocks and non-landslide blocks in the original image using $maskList_1$ and $maskList_0$, respectively.
\subsubsection{Teacher-student Network}
Inspired by the self-distillation framework in DINO \cite{46}, we developed a teacher-student network architecture that shares \textcolor{blue}{the} structure but not \textcolor{blue}{the} parameters, utilizing self-distillation for masked feature modeling and semantic feature contrast enhancement task. The encoders in both the teacher and student networks have the same structure as the feature extraction module. During the self-distillation process, the teacher network guides the student network. For the student network, we use stochastic gradient descent and backpropagation to update parameters $\theta_{s}$, while the teacher network is updated \textcolor{blue}{based on} past iterations of the student network, and its gradients are frozen. We use an exponential moving average (EMA), known as a momentum encoder, to build the parameters $\theta_{t}$ of the teacher network. The update strategy is as follows, 
\begin{equation}
\theta_{t}=\lambda\theta_{t}+\left(1- \lambda\right)\theta_{s},
\end{equation}
where $\theta_{s}$ represents the parameters of the student network, $\theta_{t}$ represents the parameters of the teacher network, and $\lambda$ is a hyperparameter set to $0.996$.

\subsubsection{Masked Feature Modeling}
We randomly select two different landslide samples to create an input pair and perform masking on each sample in the pair. The masked sample $mask\in \mathbb{R} ^{H\times W\times 3}$ is input into the student network to obtain the feature map \textcolor{blue}{$f_{mask} \in \mathbb{R} ^{\frac{H}{8} \times \frac{H}{8} \times C}$} of the masked sample; the original sample $img\in \mathbb{R} ^{H\times W\times 3}$ is input into the teacher network to obtain the feature map \textcolor{blue}{$f_{img} \in \mathbb{R} ^{\frac{H}{8} \times \frac{H}{8} \times C}$} of the original sample. Each feature point in the feature map corresponds to an $8\times 8$ pixel block in the original sample. We then flatten the feature maps $f_{mask}$ and $f_{img}$ \textcolor{blue}{along} the spatial dimension.

According to the mask position information recorded in $maskList$, we collect \textcolor{blue}{the} reconstructed feature points ${f_{mask\_i}}^{1\times 1\times C} \in f_{mask}$ from the feature map of the masked sample to construct the reconstruction features ${f_{re}}'  \in \mathbb{R} ^{N\times C}$, and collect \textcolor{blue}{the} feature points ${f_{img\_i}}^{1\times 1\times C} \in f_{img}$ corresponding to the positions from the feature map of the original sample to construct the label features ${f_{l}}'  \in \mathbb{R} ^{N\times C}$. \textcolor{blue}{Here,} $N$ corresponds to the number of blocks randomly masked in the original sample. Finally, we calculate the loss between the reconstruction features and label features using the following mean squared error (MSE) loss function,  
\begin{equation}
L_{MSE} = \frac{1}{n} \sum \left ( {f_{re}}'- {f_{l}}' \right )^{2},
\end{equation}
where $n$ represents the number of samples in the dataset, ${f_{re}}'$ and ${f_{l}}'$ denote the reconstruction features and the label features, respectively.

For the two samples in the input pair, we calculate the feature reconstruction loss for each sample and its mask, respectively, and then sum them to obtain the masked feature modeling loss $L_{Reconst}$, 
\begin{equation}
L_{Reconst} =  L_{MSE\_1} +  L_{MSE\_2},
\end{equation}
where $L_{MSE\_1}$ and $L_{MSE\_2}$ denote the feature reconstruction \textcolor{blue}{losses} for the samples in the input pair, respectively.

By masking certain landslide edge blocks and non-landslide blocks, and reconstructing the masked areas using other landslide edge, landslide interior, and non-landslide parts from the original image, we specifically enable the model to learn features of landslide edges and the background, thereby enhancing the capability to extract features of model. We set the reconstruction task at the feature map level, restoring feature points. This is because, after multiple layers of convolution, the high-level feature maps contain richer and more abstract semantic features compared to the highly similar image features in RGB images. This allows the model to acquire more knowledge.
\subsubsection{Semantic Feature Contrast Enhancement} 
We normalize and flatten the feature map $f_{mask}$ of the masked sample and the feature map $f_{img}$ of the original sample. Each feature point in the feature map corresponds to the semantic feature of an $8\times8$ pixel block in the sample. Based on the mask position information recorded in the mask module, we extract \textcolor{blue}{the} reconstructed feature points ${f_{mask_1\_i}}^{1\times 1\times C} \in f_{mask_1}$ from $mask_1$ and original feature points ${f_{img_2\_i}}^{1\times 1\times C} \in f_{img_2}$ from $img_2$, as well as reconstructed feature points ${f_{mask_2\_i}}^{1\times 1\times C} \in f_{mask_2}$ from $mask_2$ and original feature points ${f_{img_1\_i}}^{1\times 1\times C} \in f_{img_1}$ from $img_1$, and perform cross-contrast enhancement \textcolor{blue}{on} these extracted feature points, respectively. \textcolor{blue}{It is worth noting that by using two random and distinct samples for cross-contrastive learning, the feature diversity between different samples can be effectively leveraged. This allows for the construction of more positive and negative pairs, helping the model learn broader patterns and features, and enhancing its robustness.}

In cross-contrast enhancement, we label feature points corresponding to all landslide edge blocks as class-1, creating a class-1 set $\left \{ x_{k} ,y_{k}  \right \} _{k= 1,2,\dots ,N\mid y_{k} = 1} $. Feature points corresponding to all non-landslide blocks are labeled as class-0, forming a class-0 set $\left \{ \tilde{x} _{k} ,\tilde{y} _{k}  \right \} _{k= 1,2,\dots ,N\mid \tilde{y} _{k} = 0}$. We then calculate \textcolor{blue}{the} loss using a supervised contrastive learning loss function, 
\begin{equation}
L_{cons}=\sum_{i}L_{i}^{sup},
\end{equation}
\begin{scriptsize}
\begin{equation}
L_{i}^{sup} =\frac{-1}{N}\sum_{j=1}^{N}\mathrm {I}_{\left [ i\ne j \right ] } \cdot \log{\frac{\exp \left ( x_{i} \cdot x_{j}/\tau  \right ) }{\exp \left ( x_{i} \cdot x_{j}/\tau \right ) +   {\textstyle \sum_{k=1}^{N}\exp \left ( x_{i} \cdot \tilde{x}_{k}  /\tau \right ) } } }, 
\end{equation}
\end{scriptsize}
where $\mathrm{I}_{\left[i\ne j\right]}$ is the \textcolor{blue}{indicator} function with value of 1 for $i=j$ othewise 0 for $i \neq j$, $x$ and $\tilde{x}$ represent the high-level feature vectors for class-1 and class-0, \textcolor{blue}{respectively,} and $\tau$ is the temperature hyperparameter.

Finally, we sum the losses from the two cross-contrast enhancements to obtain the semantic feature contrast enhancement loss $L_{Cons}$, 
\begin{equation}
L_{Cons}=L_{cons\_1}+L_{cons\_2}.
\end{equation}

Even though we set the reconstruction task at the feature map level to learn more semantic information, it is still confined within a single sample. The semantic feature contrast enhancement performs contrastive learning among feature points from different samples, effectively increasing the distance between the features of landslide \textcolor{blue}{edges} and non-landslide categories in the high-dimensional semantic feature space. Using a supervised contrastive loss function, feature points with the same label are used to form positive sample pairs, better capturing the similarity of features within the same category. This module enhance the model's ability to distinguish between landslide edge features and background features, while also accelerating network convergence.

\subsection{Fusion Module and Decoder}
The segmentation branch extracts \textcolor{blue}{the} semantic features of the relic landslide $f_1$ and $f_2$, and the feature enhancement branch extracts the enhanced features $f_{ad\_1}$ and $f_{ad\_2}$. The enhanced features $f_{ad\_1}$ and $f_{ad\_2}$ are added point-to-point to the semantic features $f_1$ and $f_2$.

We use the same decoder proposed in our previous work \cite{44}. As shown in Fig. \ref{fig_1}, two transposed convolution layers are stacked to recover the resolution and minimize information redundancy. In addition, the dropout layer is employed to prevent  overfitting. The batch normalization layer is used to restrict the data fluctuation range, while the ReLU activation function is employed to increase network sparsity and prevent gradient vanishing. 

The fused feature is input into the decoder to perform the segmentation task. The segmentation loss is calculated using a cross-entropy loss function,
\begin{equation}
L_{CE}=-y_{i}\log{\tilde{y}_{i}}+\left( 1-y_{i}\right)\log{\left( 1-\tilde{y}_{i}\right)},
\end{equation}
where $y_{i}$ \textcolor{blue}{denotes} the ground truth and $\tilde{y}_{i}$ denotes the predicted output.

During training, we minimize three losses. The first is the masked feature modeling loss $L_{Reconst}$ within sample, the second is the semantic feature contrast loss $L_{Cons}$ between samples, and the third is the cross-entropy loss $L_{CE}$ for semantic segmentation,
\begin{equation}
Loss=\alpha L_{Reconst}+\beta L_{Cons}+\gamma L_{CE}. 
\end{equation}
Based on multi-task training, the feature enhancement branch complements and strengthens the semantic segmentation task with its features, while the segmentation branch, in turn, guides and constrains the local feature enhancement task. During testing, we add the enhanced features obtained from the teacher network to the semantic features extracted by the feature extraction module, and input them into the decoder to obtain the segmentation results.

\section{Experiments}
In this section, we evaluate the proposed MRIFE on a real dataset and compare it performance with our previous work.
\subsection{Experimental Dataset}
The study area is located in the northwest of China, which is in the transition zone between the western Qinling Mountains and the Longxi Loess Plateau. The soil parent material in this area is eluvial and accumulative, leading to soil with weak water permeability and erosion resistance. Sparse vegetation and heavy rainfall contribute to poor geological stability, which is conducive to the occurrence of landslides.

\begin{figure}[ht]
\centering
\includegraphics[scale=0.35]{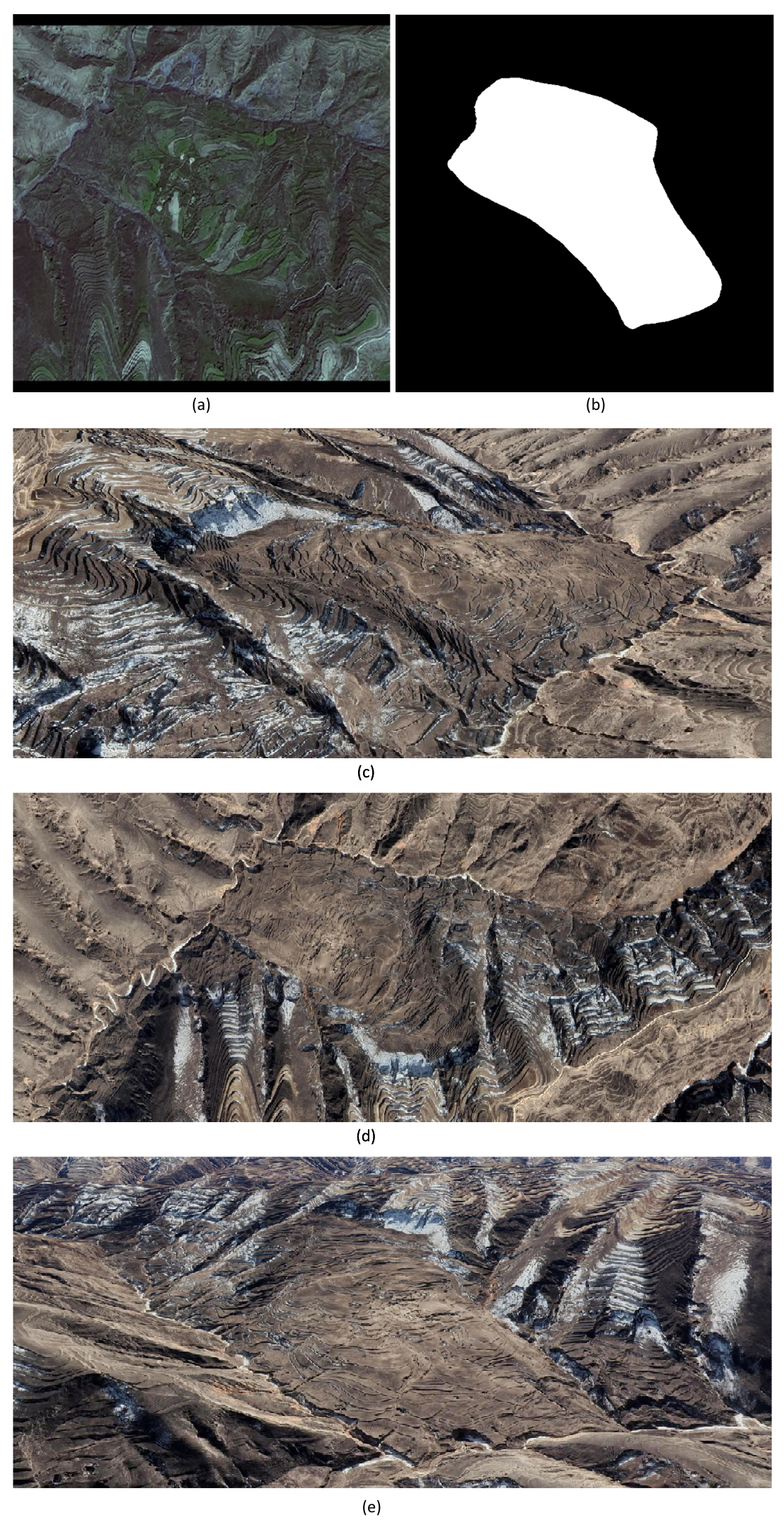}
\caption{(a) Landslide similar to the background; (b) label; (c)-(e) side view/top vie/front view of the landslide in Google Earth.}
\label{fig_3}
\end{figure}

The landslides in the study area are ancient and currently stable. After long periods of evolution and geological activity, their shapes, colors, and textures have become very similar to the surrounding environment. This similarity makes key visual indicators such as the landslide back walls and side walls very ambiguous. Furthermore, the main bodies of some landslides are obscured by farmland or residential areas, making landslide detection much more challenging. The employed dataset is constructed by a professional institution \footnote{ Institute of Remote Sensing and Digital Earth, Chinese Academy of Sciences}, and the landslide samples are labeled by experts following the following the procedure outlined below: 1) First, the potential landslides are searched according to the morphological characteristics from DEM and HRSI data in the rupture zones along the banks of the river valley; 2) Next, each potential landslide is distinguished from the three profiles of high-resolution 3D images on Google Earth; 3) Finally, the boundary of confirmed landslide is labeled on the 2D HRSI image.  Fig. \ref{fig_3} shows a hard sample.

\subsection{Data Preprocessing}
Images containing landslides are cropped from the original HRSIs and annotated to create positive samples, while images without landslides are randomly cropped to generate negative samples. All samples are resized to $512\times512$ pixels using zero-padding and scaling techniques to ensure uniform sample size, facilitating training.

Samples are preprocessed using histogram equalization and data augmentation techniques\textcolor{blue}{, such as} horizontal flipping, vertical flipping, and rotation. The dataset is divided into training, validation, and test sets in a $6: 2: 2$ ratio. \textcolor{blue}{The detailed} dataset division is shown in Table \ref{table_1}.
\begin{table}[h]
\centering
\caption{Dataset Division} 
\label{table_1}
\begin{tabular}{lccc}
\toprule
 &Training &Validation &Test\\
\midrule
Slide   &1080 &360 &60\\
Non-slide   &360 &120 &20\\
\bottomrule
\end{tabular}
\end{table}

\subsection{Performance Metrics}
Pixel-level semantic segmentation metrics are used, \textcolor{blue}{including} pixel accuracy (PA), precision, recall, F1-score, mean intersection over union (mIoU), and landslide intersection over union (1-IoU).

\begin{small}
\begin{equation}
PA=\frac{TP+TN}{TP+FP+TN+FN},
\end{equation}
\begin{equation}
precision=\frac{TP}{TP+FP},  
\end{equation}
\begin{equation}
recall=\frac{TP}{TP+FN},
\end{equation}
\begin{equation}
1-mIoU=\frac{TP}{TP+FP+FN},
\end{equation}
\begin{equation}
mIoU=\frac{1}{2}\times\left( \frac{TP}{TP+FP+FN}+\frac{TN}{TN+FN+FP}\right),
\end{equation}
\begin{equation}
F1-score=2\times\frac{precision\times recall}{precision+recall},
\end{equation}
\end{small}
where $TP$, $TN$, $FP$ and $FN$ represent the true positive, true negative, false positive, and false negative at the pixel level, respectively.

\subsection{Reference Models}
Although several landslide detection models \cite{50,51} and HRSI-based semantic segmentation models \cite{52} have been published, unfortunately, we cannot directly compare our model with them for three reasons: 1) They utilize the aerial images, DEM, or SAR, which differ from HRSI data; 2) Their detection targets are not landslides; 3) They have not made their model source code publicly available, making it impossible to replicate their results. Therefore, we must validate the advantages of our proposed model through comparisons with baseline models and ablation experiments.

We choose Deeplabv3+ \cite{27} as the baseline model, and \textcolor{blue}{design} three comparative experiments: 1) The baseline model Deeplabv3+; 2) The ICSSN from our previous work \cite{44}; 3) The proposed MRIFE. For all comparison models, stochastic gradient descent is selected as the optimizer, with a learning rate of $0.007$, momentum of $0.9$, and weight decay of $0.0005$. Table \ref{table_2} lists the hyperparameters used for model training.

\begin{table}[h]
\centering
\caption{Training Hyperparameters} %Quantitative comparison results
\label{table_2}
\begin{tabular}{lc}
\toprule
\multicolumn{1}{c}{Hyperparameter} & Value\\
\midrule
Number of workers & 4\\
Batch size & 4\\
Optimizer & SGD\\
Momentum & 0.9\\
learning rate & 0.007\\
Weight decay & 0.0005\\
Epoch &100\\
\bottomrule
\end{tabular}
\end{table}

All experiments are conducted using PyTorch on two Nvidia 3090 GPUs, a 12th-generation Intel(R) Core(TM) i9-12900K, and 128 GB of memory.
\begin{table*}[b]
\centering
\caption{Quantitative Comparison Results on Segmentation Task} 
\label{table_3}
\begin{tabular}{lcccccc}
\toprule
\multicolumn{1}{c}{Method} & PA & Precision & Recall & 1-IoU & mIoU & F1-score\\
\midrule
Baseline & 0.9445 & 0.4226 & 0.6284 & 0.3381 & 0.6405 & 0.5054 \\
ICSSN & 0.9493 & 0.4531 & 0.5911 & 0.3743 & 0.6610 & 0.5446 \\
\textbf{MRIFE} & 0.9447 & \textbf{0.5347} & 0.5981 & \textbf{0.3975} & \textbf{0.6680} & \textbf{0.5646} \\
\bottomrule
\end{tabular}
\end{table*}

\begin{figure*}[h]
\centering
\begin{tabular}{c@{\hspace{2pt}}c@{\hspace{2pt}}c@{\hspace{2pt}}c@{\hspace{2pt}}c@{\hspace{4pt}}c@{\hspace{2pt}}c@{\hspace{2pt}}c@{\hspace{2pt}}c@{\hspace{2pt}}c}
%第1行
\includegraphics[width=0.08\textwidth]{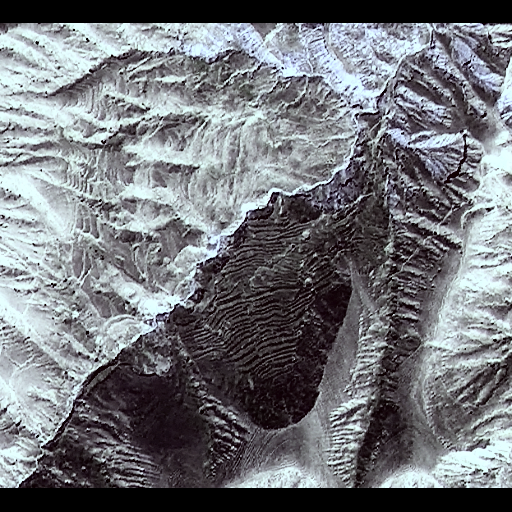} & 
\includegraphics[width=0.08\textwidth]{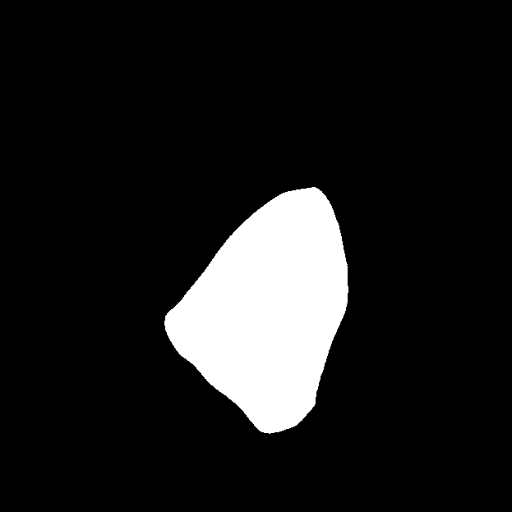} & 
\includegraphics[width=0.08\textwidth]{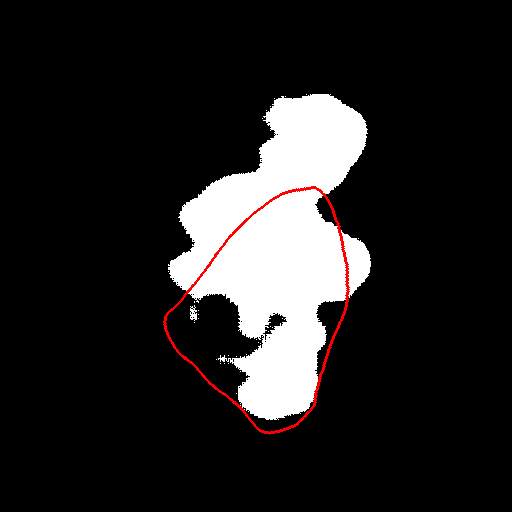} & 
\includegraphics[width=0.08\textwidth]{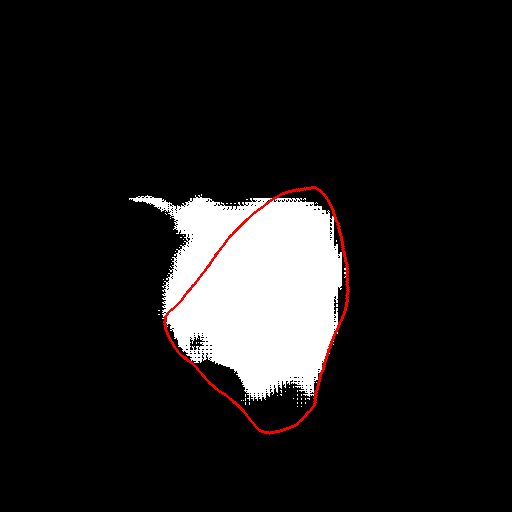} & 
\includegraphics[width=0.08\textwidth]{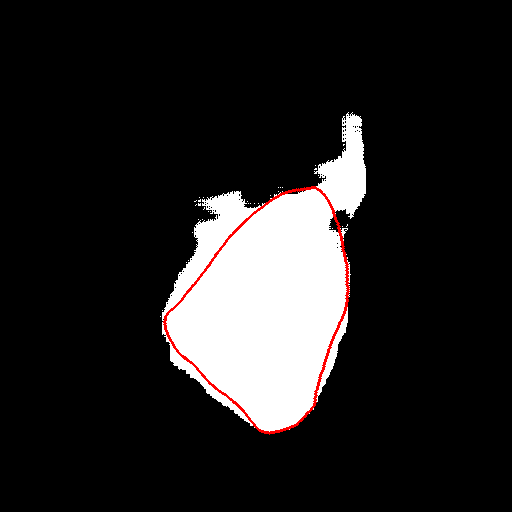} &
\includegraphics[width=0.08\textwidth]{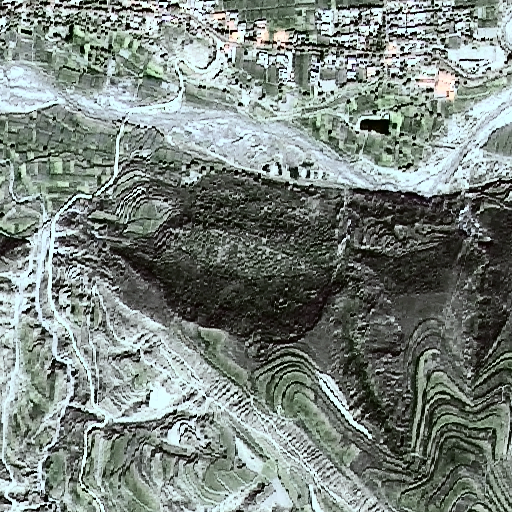} & 
\includegraphics[width=0.08\textwidth]{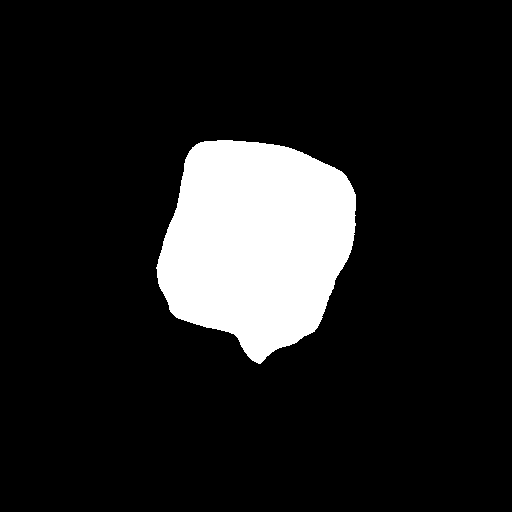} & 
\includegraphics[width=0.08\textwidth]{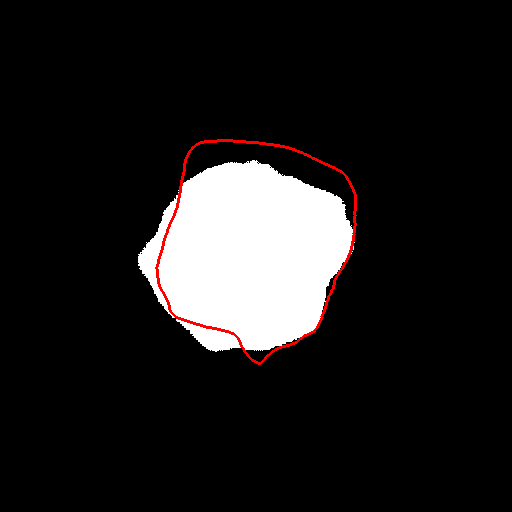} & 
\includegraphics[width=0.08\textwidth]{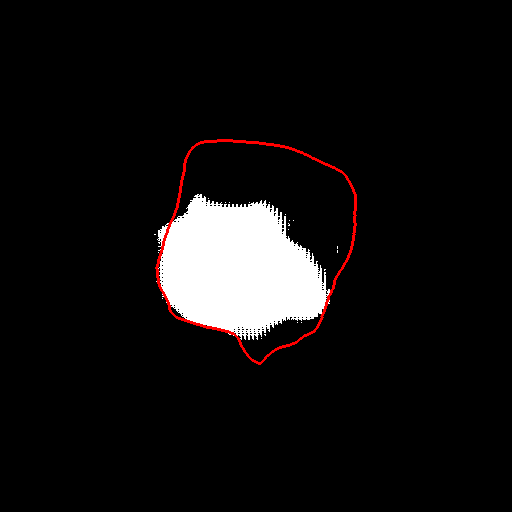} & 
\includegraphics[width=0.08\textwidth]{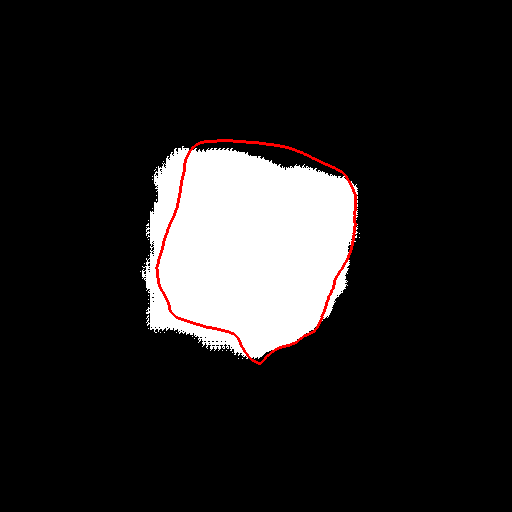} \\[-2pt]
%第2行
\includegraphics[width=0.08\textwidth]{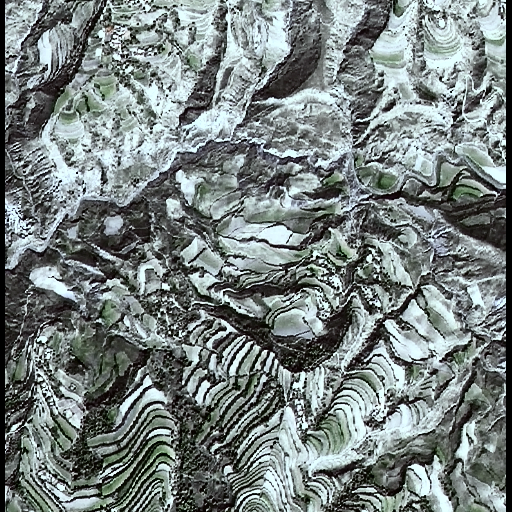} & 
\includegraphics[width=0.08\textwidth]{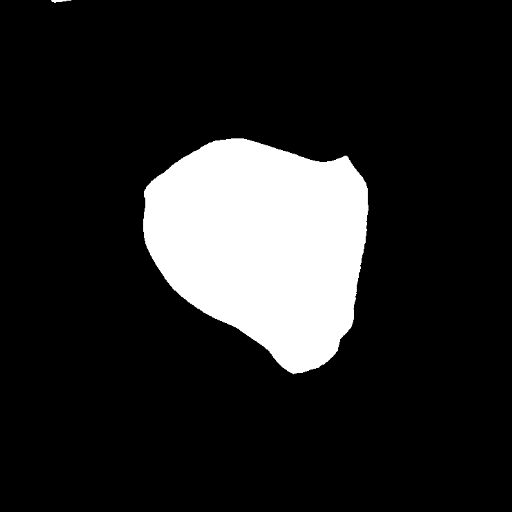} & 
\includegraphics[width=0.08\textwidth]{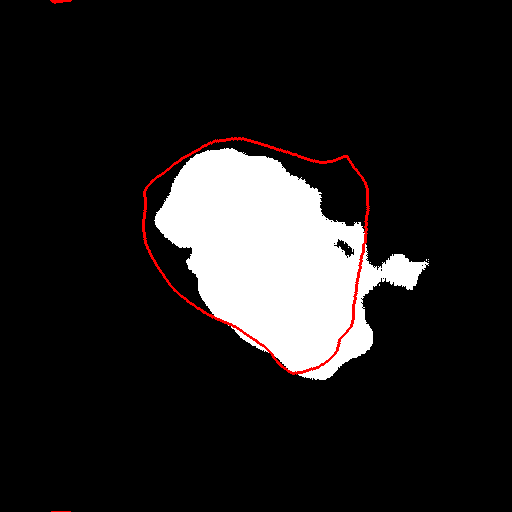} & 
\includegraphics[width=0.08\textwidth]{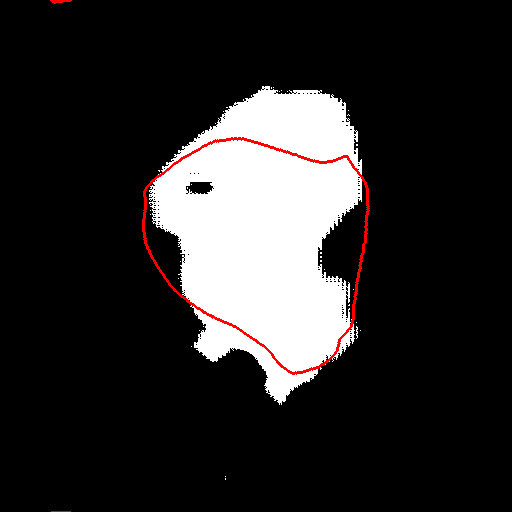} & 
\includegraphics[width=0.08\textwidth]{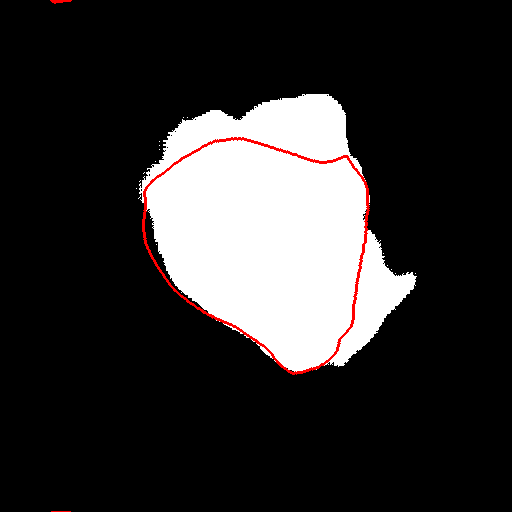} &
\includegraphics[width=0.08\textwidth]{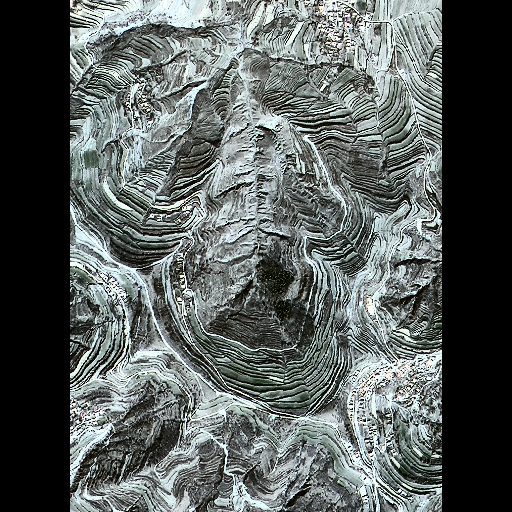} & 
\includegraphics[width=0.08\textwidth]{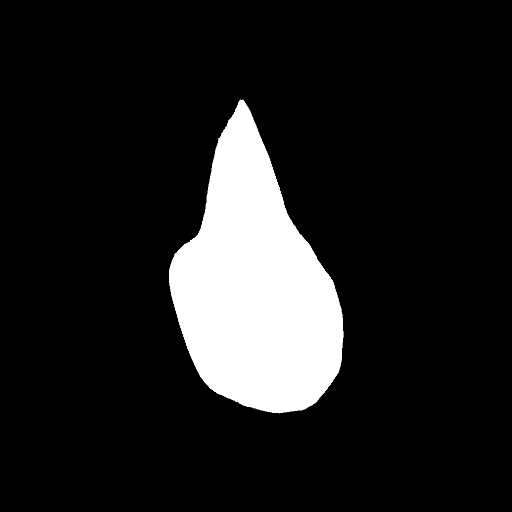} & 
\includegraphics[width=0.08\textwidth]{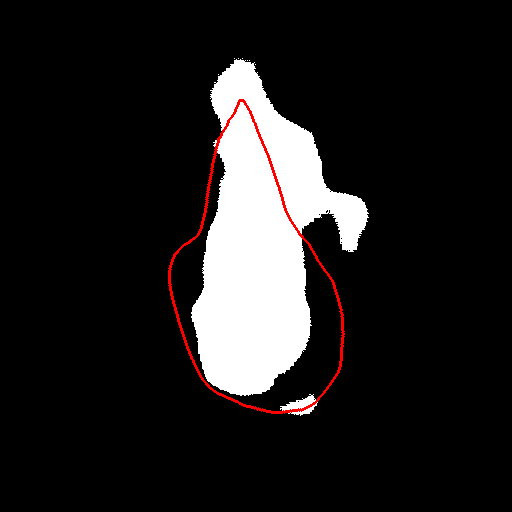} & 
\includegraphics[width=0.08\textwidth]{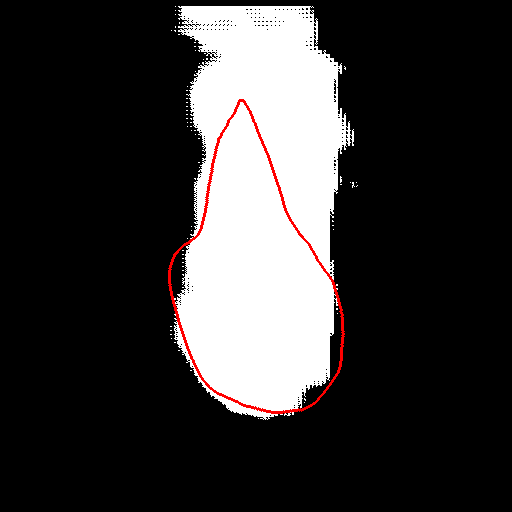} & 
\includegraphics[width=0.08\textwidth]{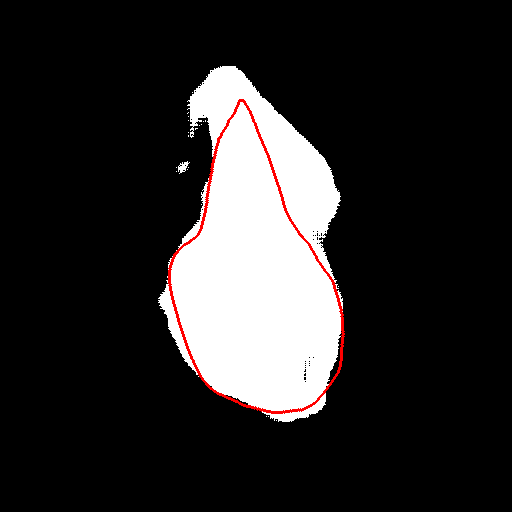} \\[-2pt]
%第3行
\includegraphics[width=0.08\textwidth]{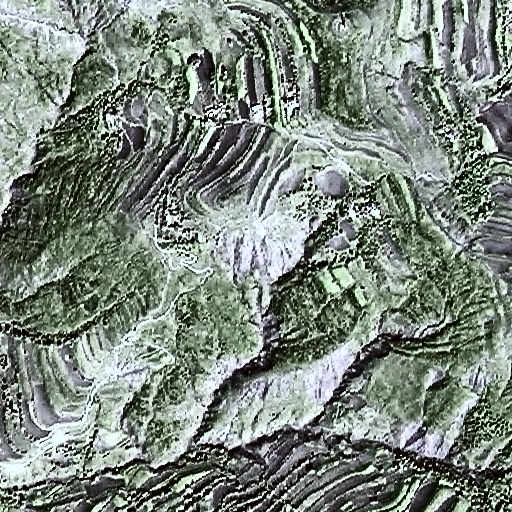} & 
\includegraphics[width=0.08\textwidth]{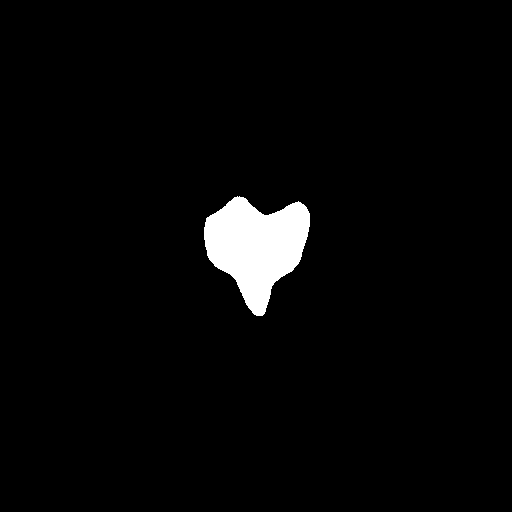} & 
\includegraphics[width=0.08\textwidth]{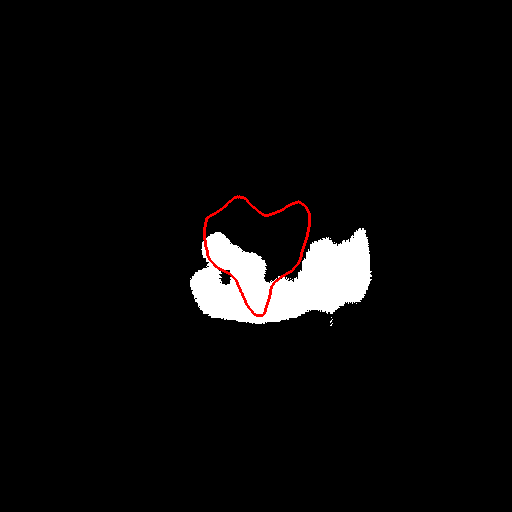} & 
\includegraphics[width=0.08\textwidth]{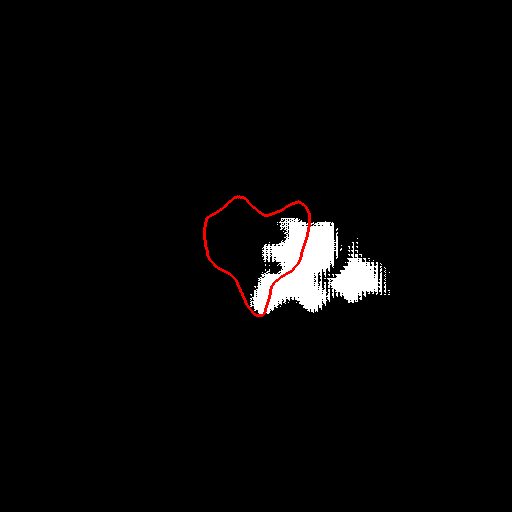} & 
\includegraphics[width=0.08\textwidth]{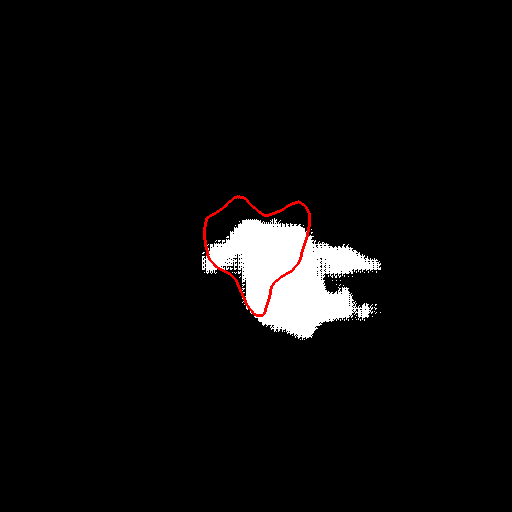} &
\includegraphics[width=0.08\textwidth]{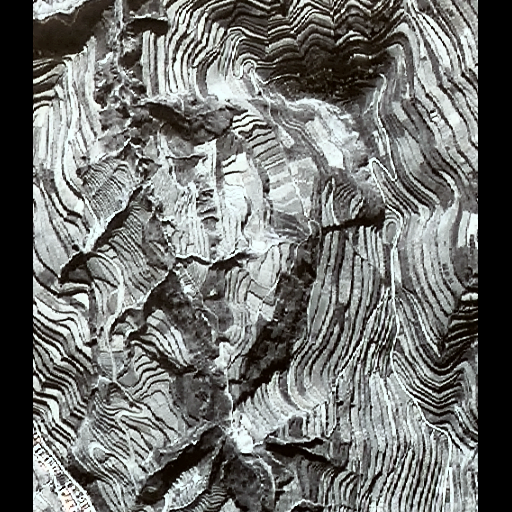} & 
\includegraphics[width=0.08\textwidth]{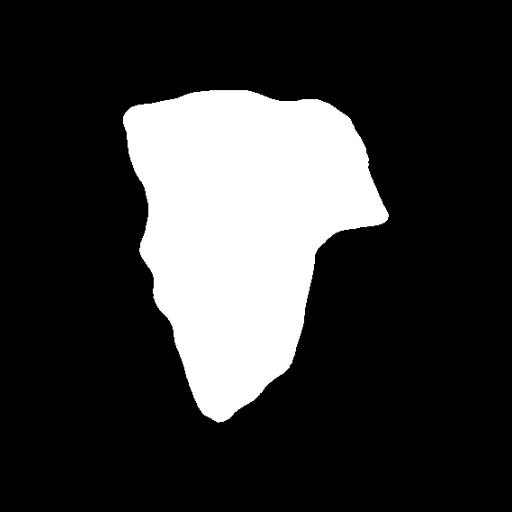} & 
\includegraphics[width=0.08\textwidth]{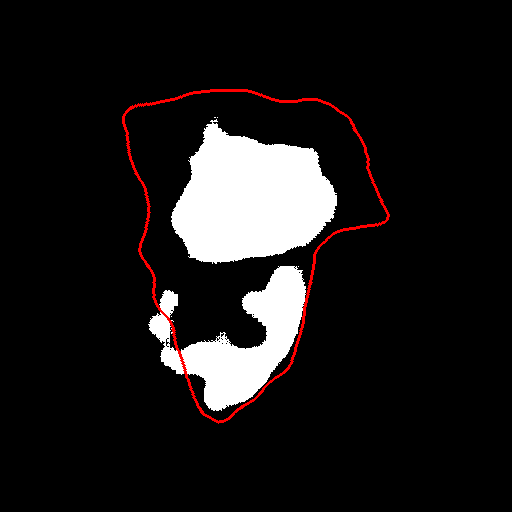} & 
\includegraphics[width=0.08\textwidth]{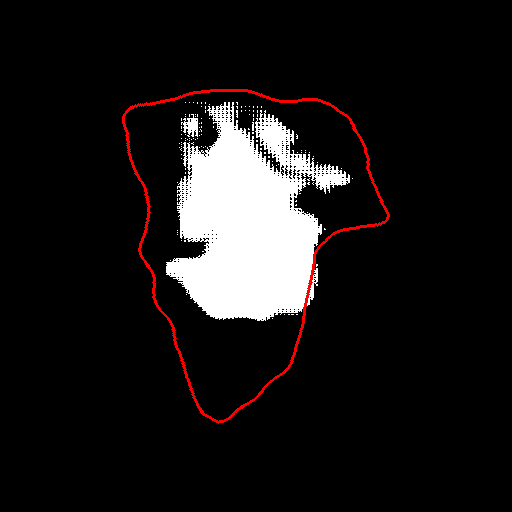} & 
\includegraphics[width=0.08\textwidth]{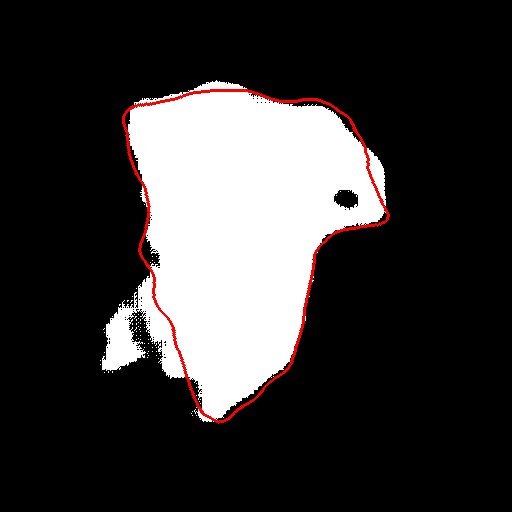} \\[-2pt]
%第4行
\includegraphics[width=0.08\textwidth]{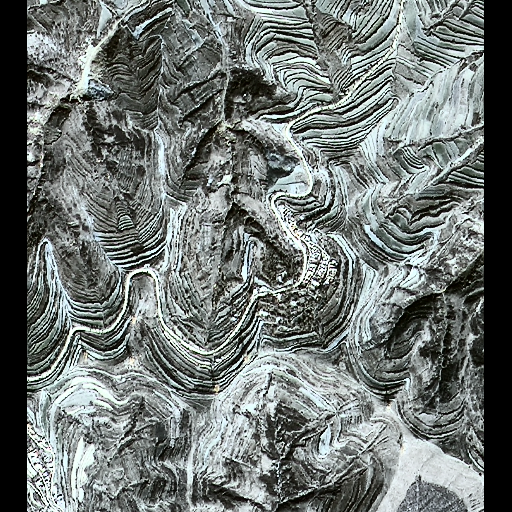} & 
\includegraphics[width=0.08\textwidth]{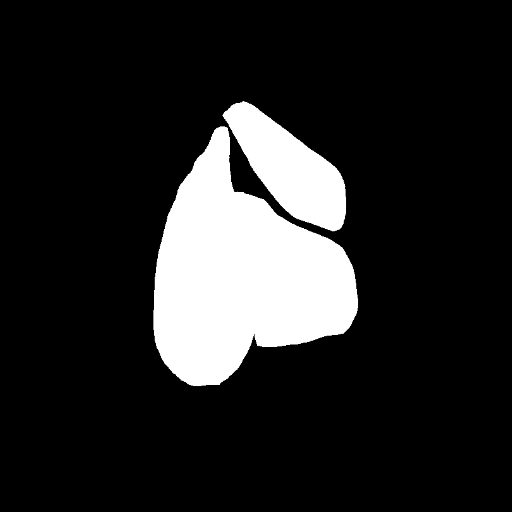} & 
\includegraphics[width=0.08\textwidth]{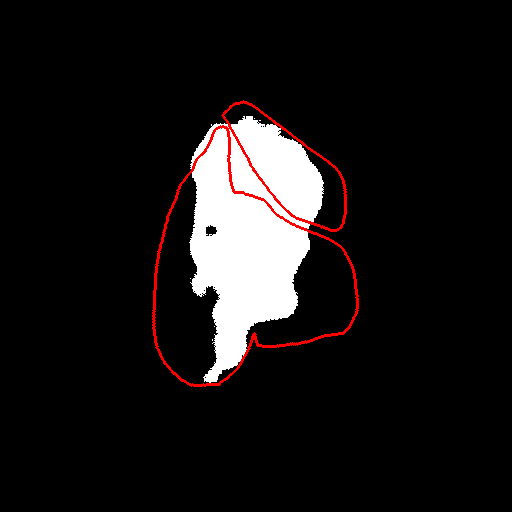} & 
\includegraphics[width=0.08\textwidth]{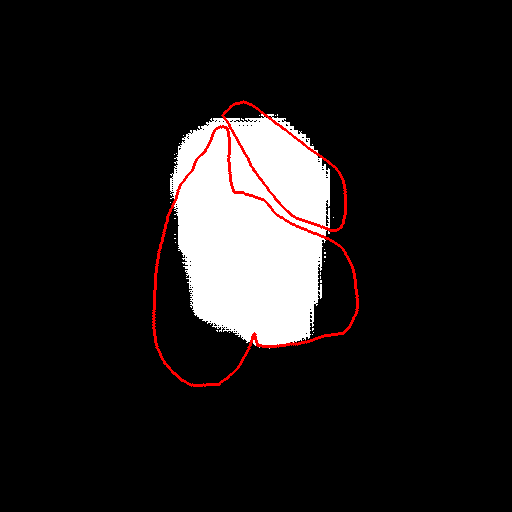} & 
\includegraphics[width=0.08\textwidth]{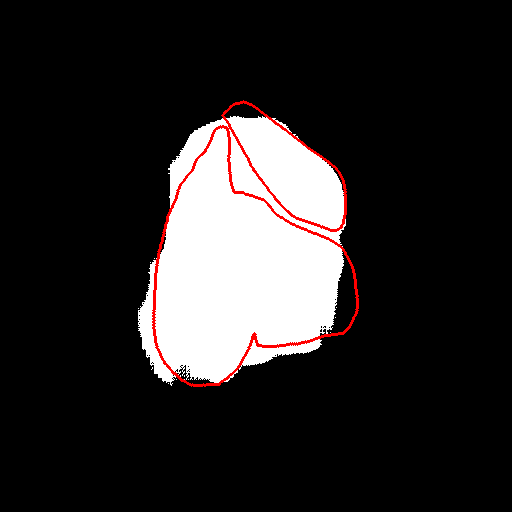} &
\includegraphics[width=0.08\textwidth]{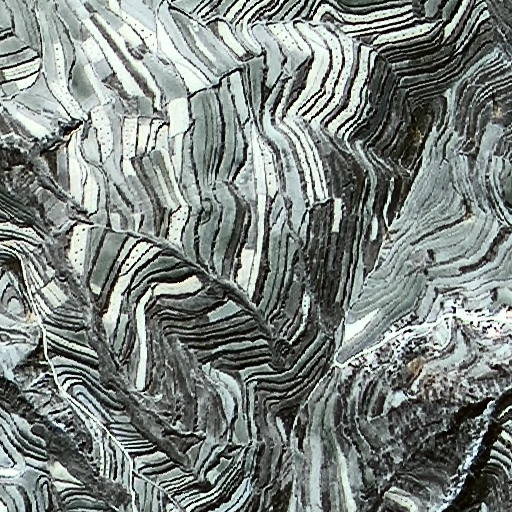} & 
\includegraphics[width=0.08\textwidth]{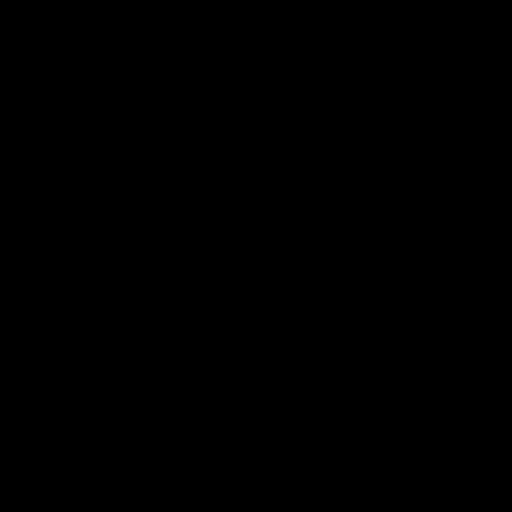} & 
\includegraphics[width=0.08\textwidth]{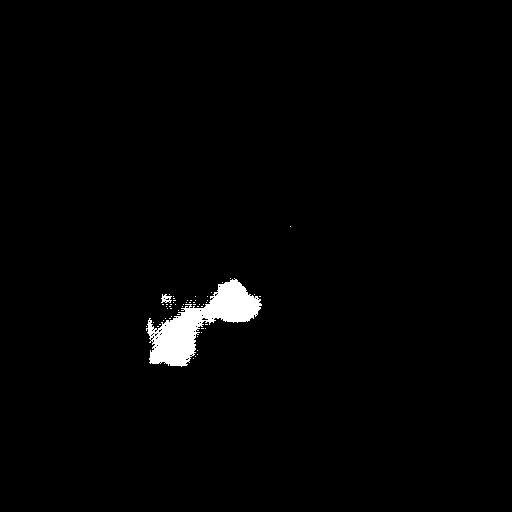} & 
\includegraphics[width=0.08\textwidth]{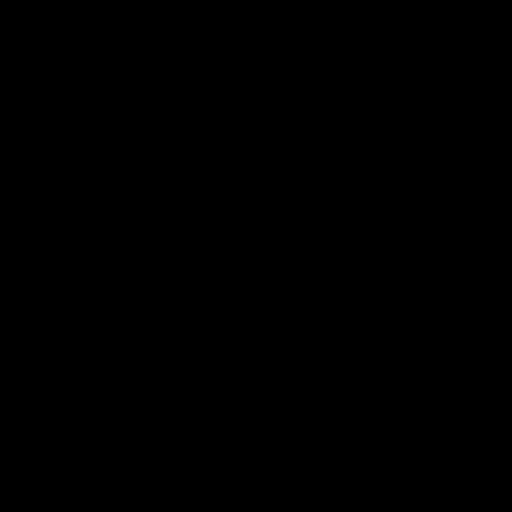} & 
\includegraphics[width=0.08\textwidth]{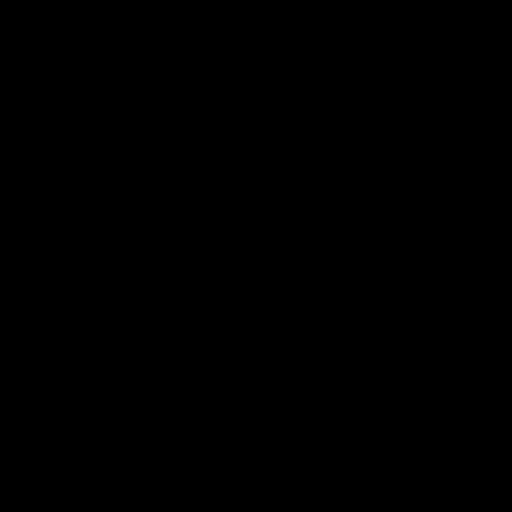} \\
(a) & (b) & (c) & (d) & (e) & (a) & (b) & (c) & (d) & (e) \\ 
\end{tabular}
\caption{Visualized results of the comparative experiments. (a) Input; (b) label; (c)-(e) prediction: baseline, ICSSN, and MRIFE. The red line indicates the boundary.}
\label{fig_4}
\end{figure*}

\section{Results and Discussions}
In this section, we present detailed results of the experiments, including both numerical and visualized results.

\subsection{Comparative Experiments}
In this subsection, we present and analyze the quantitative comparison results of the segmentation task obtained by the three comparison models on \textcolor{blue}{the} same relic landslide dataset.

\begin{figure*}[t]
\centering
\begin{tabular}{c@{\hspace{2pt}}c@{\hspace{2pt}}c@{\hspace{2pt}}c@{\hspace{2pt}}c@{\hspace{4pt}}c@{\hspace{2pt}}c@{\hspace{2pt}}c@{\hspace{2pt}}c@{\hspace{2pt}}c}
%第1行
\includegraphics[width=0.08\textwidth]{lunwen/result/img/0_gh_69.png} & 
\includegraphics[width=0.08\textwidth]{lunwen/result/label/0_gh_69.png} & 
\includegraphics[width=0.08\textwidth]{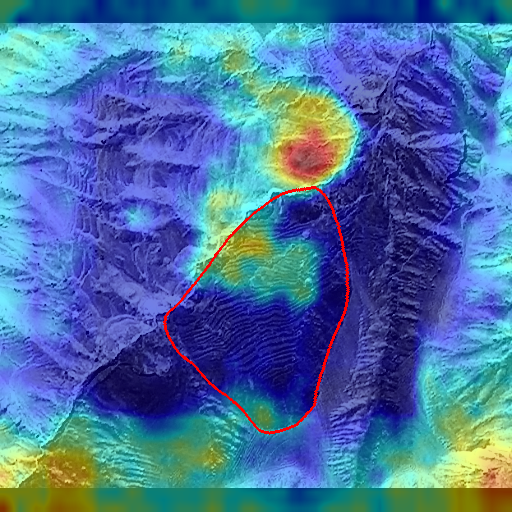} & 
\includegraphics[width=0.08\textwidth]{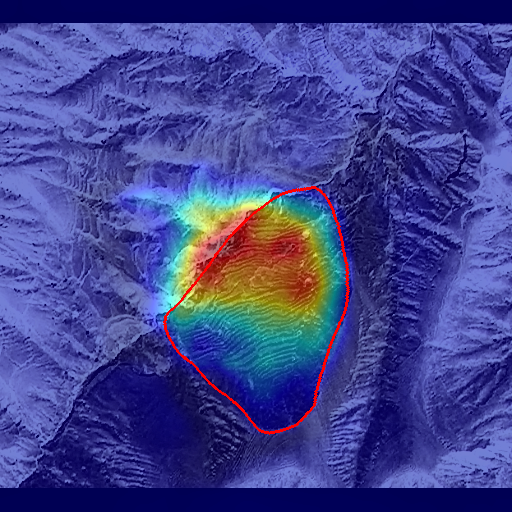} & 
\includegraphics[width=0.08\textwidth]{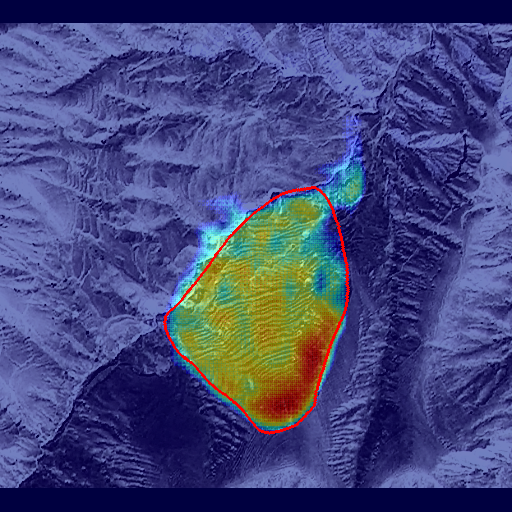} &
\includegraphics[width=0.08\textwidth]{lunwen/result/img/12_ltao_152168.png} & 
\includegraphics[width=0.08\textwidth]{lunwen/result/label/12_ltao_152168.png} & 
\includegraphics[width=0.08\textwidth]{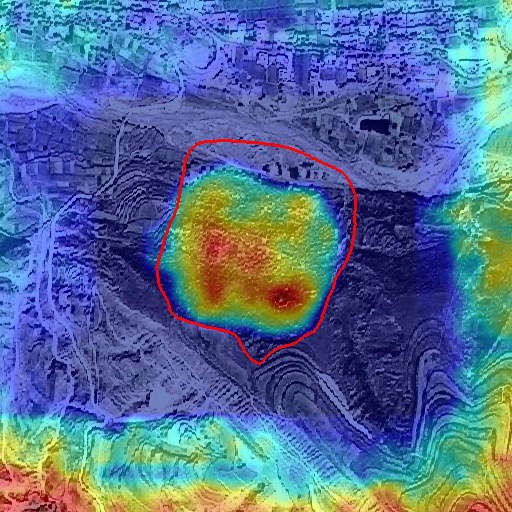} & 
\includegraphics[width=0.08\textwidth]{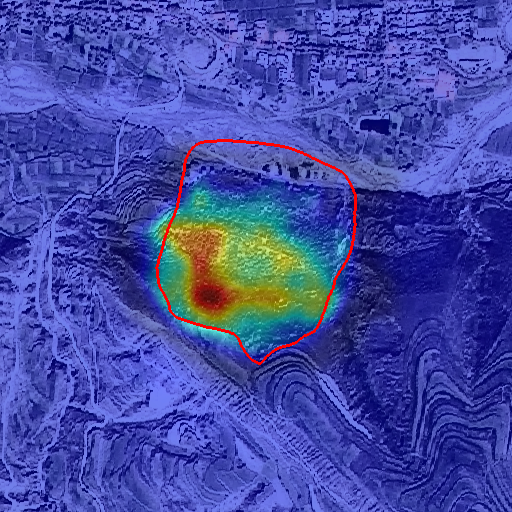} & 
\includegraphics[width=0.08\textwidth]{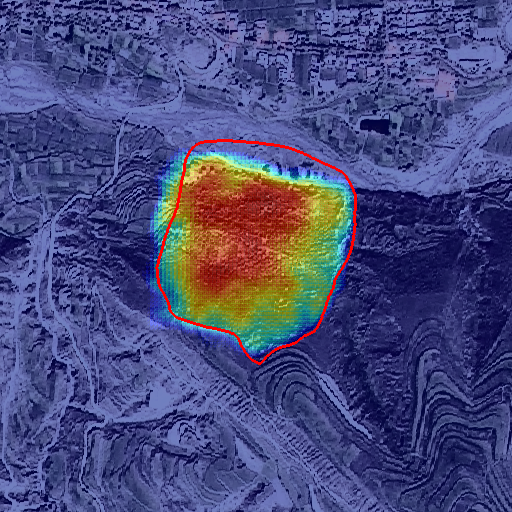} \\[-2pt]
%第2行
\includegraphics[width=0.08\textwidth]{lunwen/result/img/1_ltao_137.png} & 
\includegraphics[width=0.08\textwidth]{lunwen/result/label/1_ltao_137.png} & 
\includegraphics[width=0.08\textwidth]{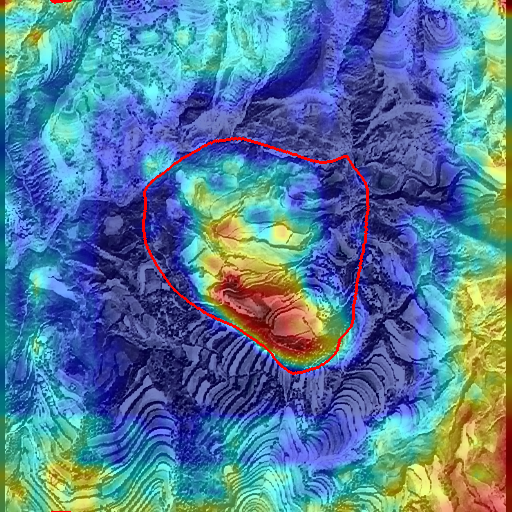} & 
\includegraphics[width=0.08\textwidth]{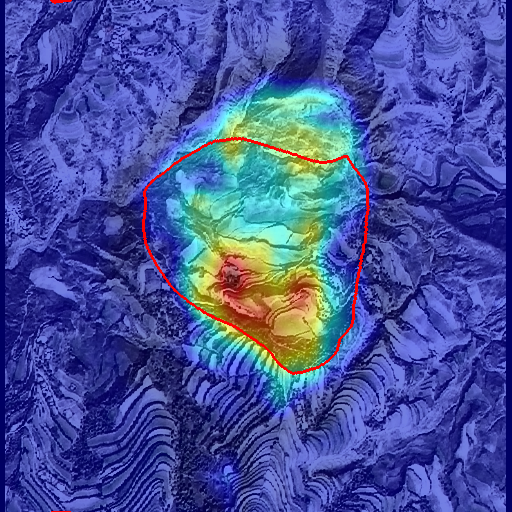} & 
\includegraphics[width=0.08\textwidth]{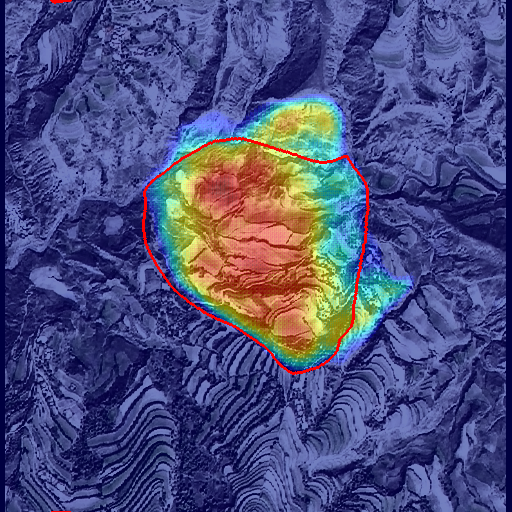} &
\includegraphics[width=0.08\textwidth]{lunwen/result/img/16_z05.png} & 
\includegraphics[width=0.08\textwidth]{lunwen/result/label/16_z05.png} & 
\includegraphics[width=0.08\textwidth]{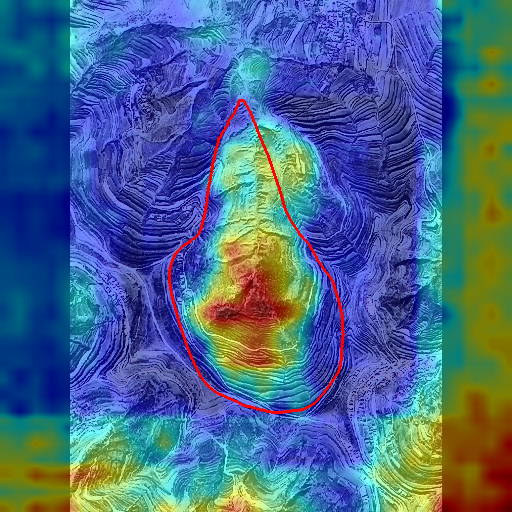} & 
\includegraphics[width=0.08\textwidth]{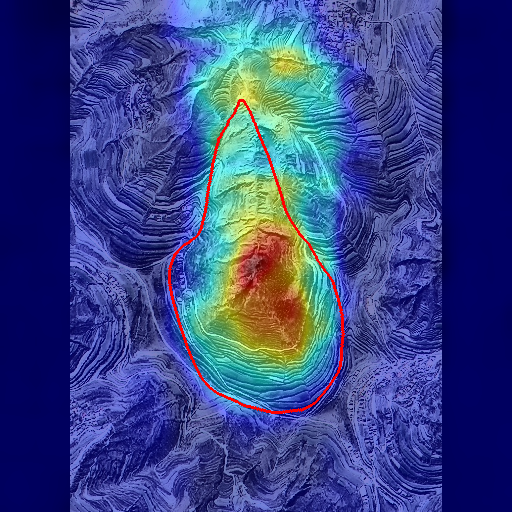} & 
\includegraphics[width=0.08\textwidth]{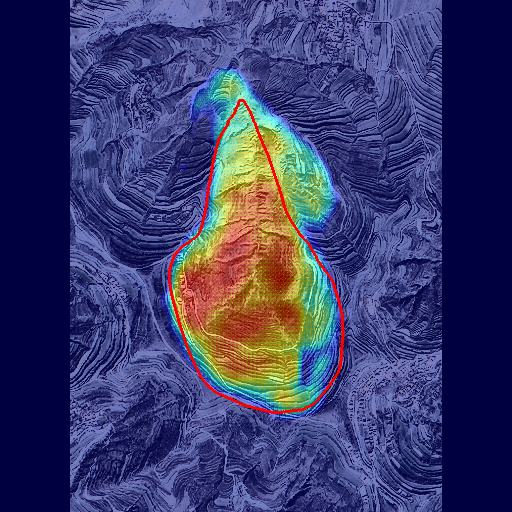} \\[-2pt]
%第3行
\includegraphics[width=0.08\textwidth]{lunwen/result/img/1_wy_7488.png} & 
\includegraphics[width=0.08\textwidth]{lunwen/result/label/1_wy_7488.png} & 
\includegraphics[width=0.08\textwidth]{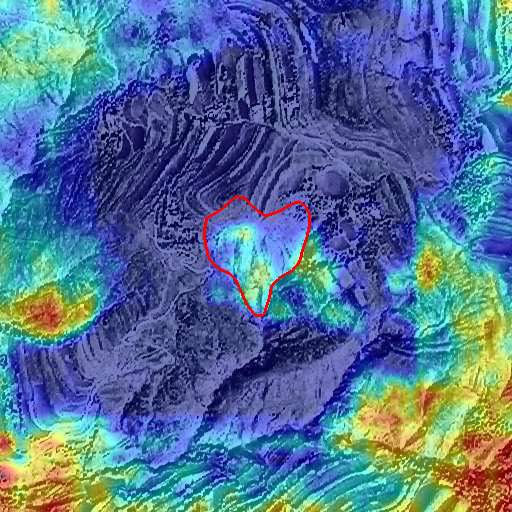} & 
\includegraphics[width=0.08\textwidth]{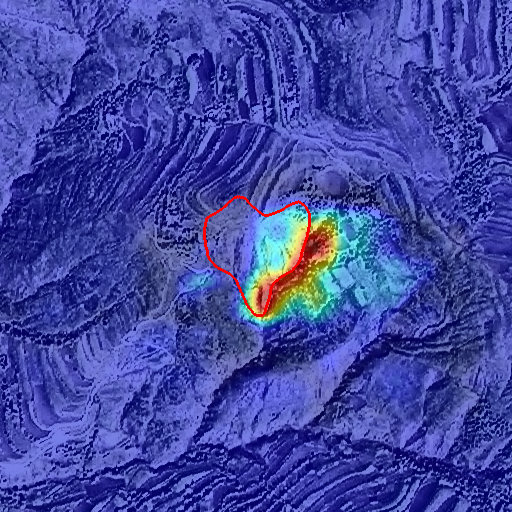} & 
\includegraphics[width=0.08\textwidth]{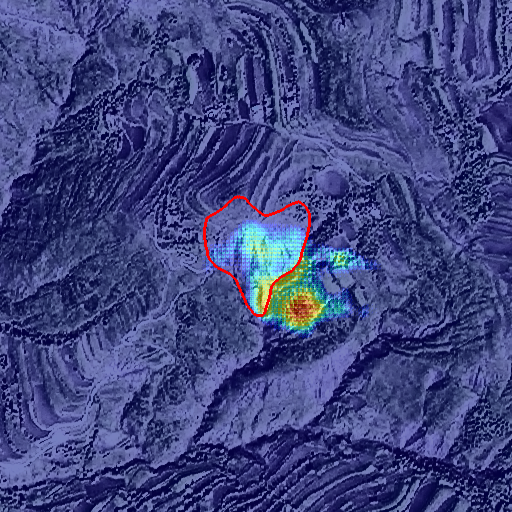} &
\includegraphics[width=0.08\textwidth]{lunwen/result/img/4_j01.png} & 
\includegraphics[width=0.08\textwidth]{lunwen/result/label/4_j01.png} & 
\includegraphics[width=0.08\textwidth]{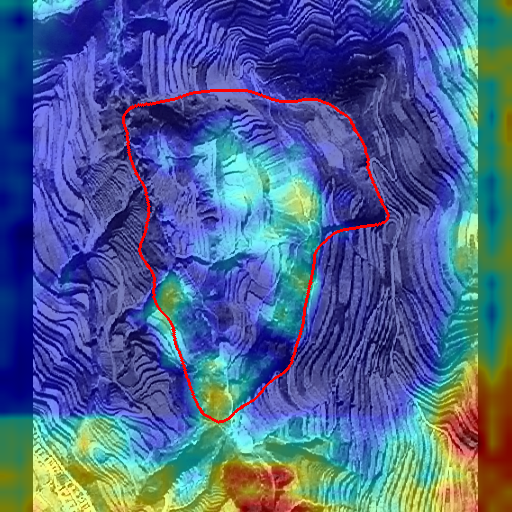} & 
\includegraphics[width=0.08\textwidth]{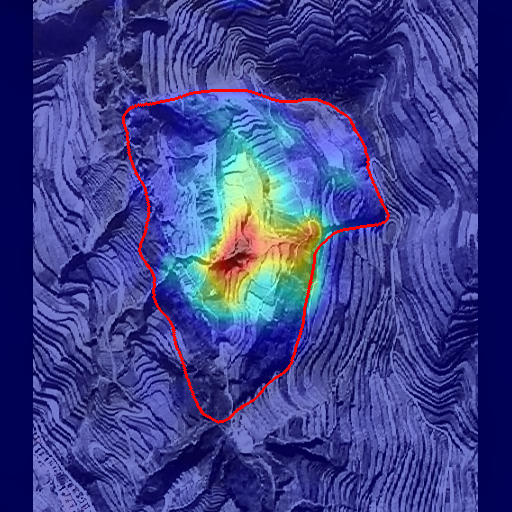} & 
\includegraphics[width=0.08\textwidth]{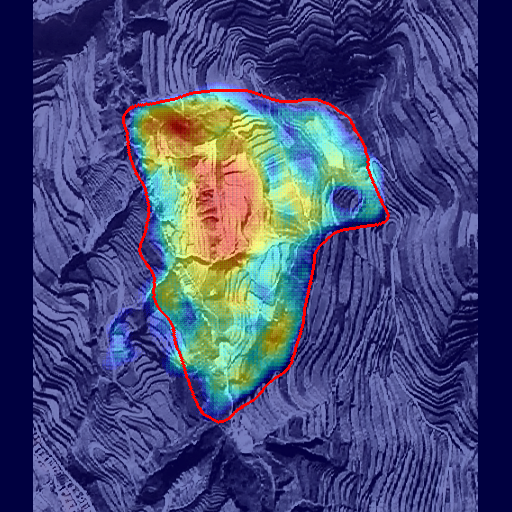} \\[-2pt]
%第4行
\includegraphics[width=0.08\textwidth]{lunwen/result/img/10_z05.png} & 
\includegraphics[width=0.08\textwidth]{lunwen/result/label/10_z05.png} & 
\includegraphics[width=0.08\textwidth]{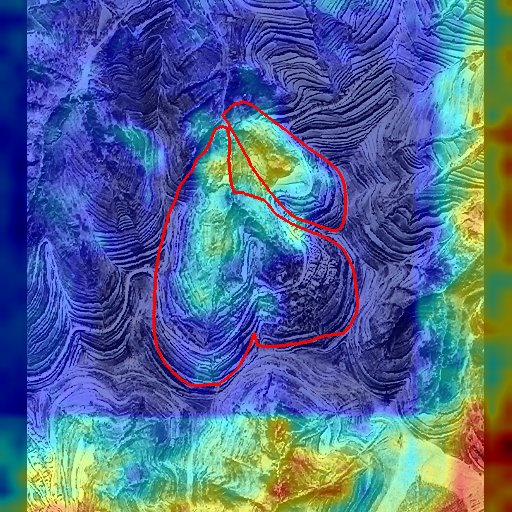} & 
\includegraphics[width=0.08\textwidth]{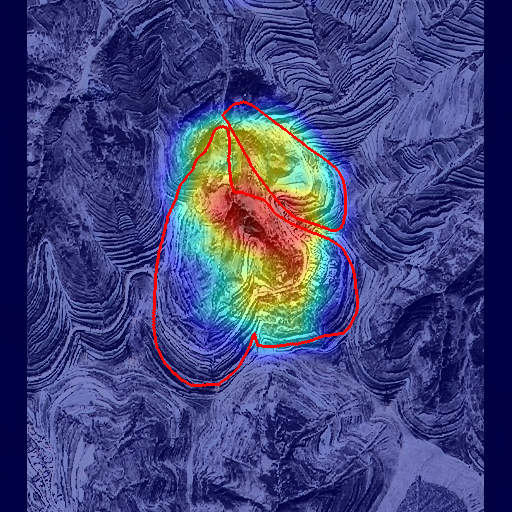} & 
\includegraphics[width=0.08\textwidth]{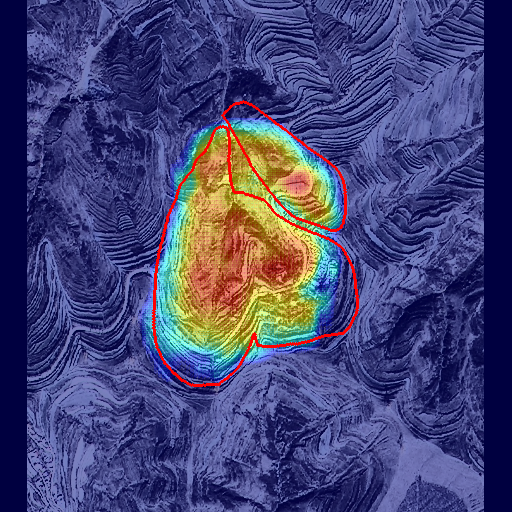} &
\includegraphics[width=0.08\textwidth]{lunwen/result/img/zjc852_v.png} & 
\includegraphics[width=0.08\textwidth]{lunwen/result/label/zjc852_v.png} & 
\includegraphics[width=0.08\textwidth]{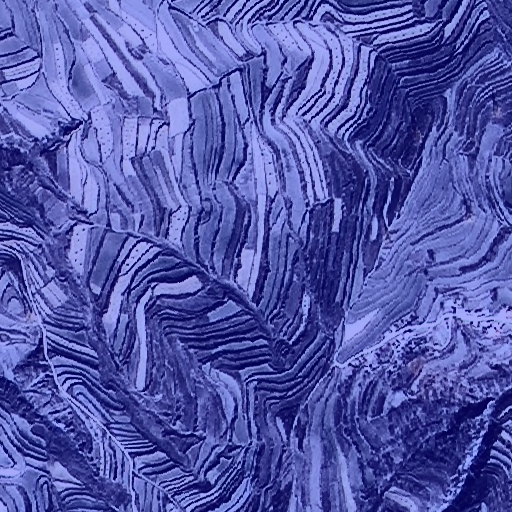} & 
\includegraphics[width=0.08\textwidth]{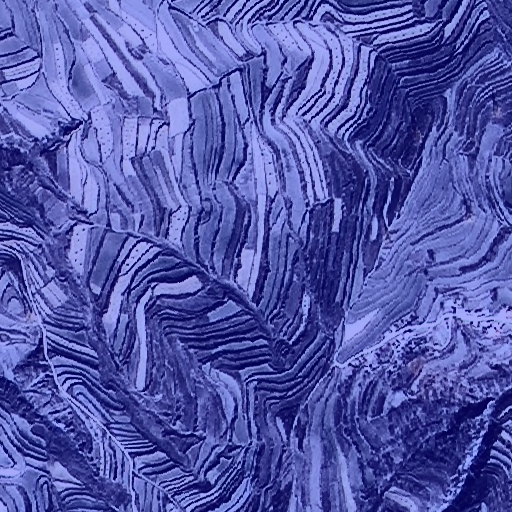} & 
\includegraphics[width=0.08\textwidth]{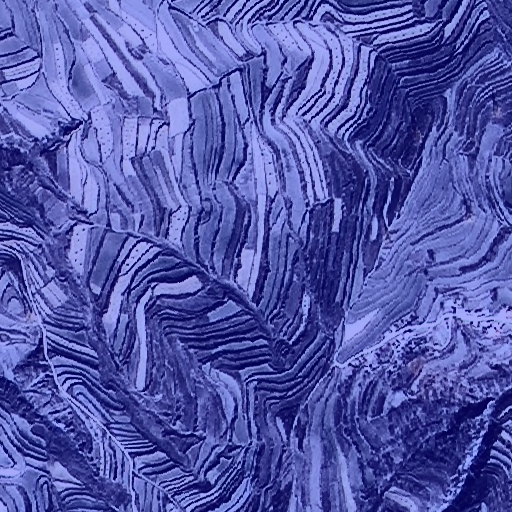} \\
(a) & (b) & (c) & (d) & (e) & (a) & (b) & (c) & (d) & (e) \\ 
\end{tabular}
\caption{Grad-CAM results of the comparative experiments. (a) Input; (b) label; (c)-(e) heat map: baseline, ICSSN, and MRIFE. Red line indicates the boundary.}
\label{fig_5}
\end{figure*}

Table \ref{table_3} lists the numerical results, from which we can make the following observations: the proposed MRIFE model achieves the best performance among all comparison models. Specifically, compared to the baseline model, the precision increases from 0.4226 to 0.5347, the mIoU increases from 0.6405 to 0.6680, the 1-IoU increases from 0.3381 to 0.3934, and the F1-score increases from 0.5054 to 0.5646. In comparison to the ICSSN model, the precision increases from 0.4531 to 0.5347, the mIoU increases from 0.6610 to 0.6680, the 1-IoU increases from 0.3743 to 0.3934, and the F1-score increases from 0.5446 to 0.5646. These improvements clearly demonstrate \textcolor{blue}{the effectiveness} of the proposed model for relic landslide detection. Fig. \ref{fig_4} shows the visualized results. It can be observed from the figure that, compared to the baseline model and the ICSSN model, the proposed MRIFE model classifies pixels more accurately, with fewer false positives and false negatives. Moreover, the predictions made by the proposed model have more precise shapes and better accuracy in segmenting target edges. This  is consistent with the numerical results.

Heat map analysis can identify features most relevant to prediction outcomes. We employ the currently popular Gradient-weighted Class Activation Mapping (Grad-CAM) algorithm \cite{53}, which visualizes the contribution of each area of the input image to the prediction results by weighting feature maps and overlaying them on the original image. We present the heat maps of key intermediate layers of the three comparison models in Fig. \ref{fig_5}. The critical areas in heat maps of the proposed MRIFE model are concentrated around the edges of landslides, like the back walls and side walls. Moreover, the heat maps display areas with high contributions to the prediction outcomes that better align with the actual target shapes compared to the baseline and the ICSSN. These heat maps further validate the effectiveness of the model.

\subsection{Ablation Experiments}
To further verify the effectiveness of the proposed MRIFE model and analyze the improvements of performance, we design ablation experiments in this subsection. The experimental results are shown in Table \ref{table_4}, where B represents the baseline model Deeplabv3+, M represents the proposed masked feature modeling, and C represents the proposed semantic feature contrast enhancement. From the table, we can make the following observations.
\begin{table}[h]
\centering
\caption{Ablation Experiments} 
\label{table_4}
\begin{tabular}{lccccc}
\toprule
 & Precision & Recall & 1-IoU & mIoU & F1-score\\
\midrule
B & 0.4226 & 0.6284 & 0.3381 & 0.6405 & 0.5054 \\
B$+$M & 0.4617 & 0.6377 & 0.3658 & 0.6552 & 0.5356 \\
B$+$M$+$C & 0.5347 & 0.5981 & 0.3975 & 0.6680 & 0.5646 \\
\midrule
\multicolumn{6}{l}{B: Baseline} \\
\multicolumn{6}{l}{M: Masked Feature Modeling} \\
\multicolumn{6}{l}{C: Semantic Feature Contrast Enhancement} \\
\bottomrule
\end{tabular}
\end{table}

% \begin{figure}[ht]
% \centering
% \includegraphics[scale=0.5]{lunwen/result/loss/B+M.png}
% \caption{Loss curves comparison of B$+$M and B$+$M$+$C.}
% \label{fig_9}
% \end{figure}

\textit{1)} When the feature enhancement branch only uses the MFM task to enhance the features extracted by the feature extraction module, in the segmentation branch, compared to the baseline model, the recall increases from $0.6284$ to $0.6377$, rising by $1\%$. The 1-IoU increases from $0.3381$ to $0.3658$, showing a $2.8\%$ improvement. The mIoU increases from $0.6405$ to $0.6552$, rising by $1.5\%$, and the F1-score performance increases from $0.5054$ to $0.5356$. The masked feature modeling task significantly enhances the model's classification accuracy for landslide pixels.

\textit{2)} When both the MFM and SFCE task are used simultaneously in the feature enhancement branch, compared to using only the MFM task, the precision from $0.4617$ to $0.5347$, rising by $7.2\%$. The 1-IoU increases from $0.3658$ to $0.3975$, showing a $2.8\%$ improvement. The mIoU increases from $0.6552$ to $0.6680$, rising by $1.3\%$, and the F1-score performance increases from $0.5356$ to $0.5646$. The SFCE task enables the model to more accurately distinguish between landslide and background pixels, reducing the probability \textcolor{blue}{of false} positive. Additionally, from the loss during training we find that \textcolor{blue}{using} only MFM, the model is prone to overfitting early in training, while the inclusion of SFCE significantly improves this situation.

\textcolor{blue}{From the numerical results, the inclusion of the MFM and SFCE tasks led to a significant improvement in the model's pixel classification accuracy and IoU metrics. Specifically, classification accuracy increased by $11\%$, and the IoU for the target class improved by $6\%$. As shown in column (e) of Figure 4, compared to column (c), the identification accuracy of landslide edges has been enhanced, further validating the effectiveness of the MFM and SFCE tasks. This improvement successfully addresses the issue of inaccurate segmentation of target and background, a challenge that previous methods struggled to overcome in visually ambiguous scenarios.}

\begin{figure}[h]
\centering
\begin{tabular}{c}
\includegraphics[width=0.4\textwidth]{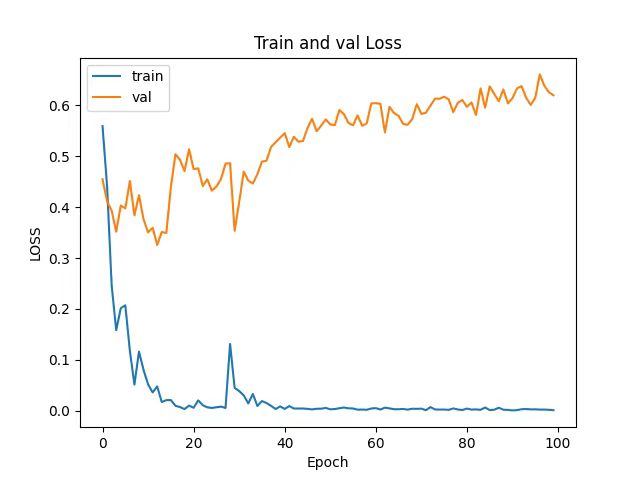} \\ 
(a) \\
\includegraphics[width=0.4\textwidth]{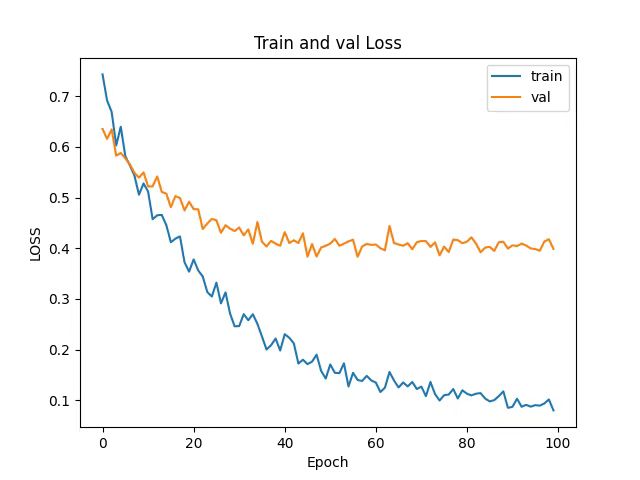} \\
(b) \\ 
\end{tabular}
\caption{Loss curve. (a) Without self-distillation framework; (b) with self-distillation framework.}
\label{fig_6}
\end{figure}

\textcolor{blue}{To evaluate the impact of the self-distillation framework on addressing the small-sized dataset problem, we conducted an ablation experiment comparing models with and without this framework. As shown in Fig. \ref{fig_6}, the training loss curve indicates that the model without the self-distillation framework began to overfit early in the training process and had difficulty converging. In contrast, the self-distillation framework leveraged feature diversity to accelerate model convergence, effectively mitigating the overfitting problem commonly encountered with the small-sized dataset}

\begin{figure}[h]
\centering
\begin{tabular}{c@{\hspace{2pt}}c@{\hspace{2pt}}c@{\hspace{2pt}}c@{\hspace{2pt}}c}
%第1行
\includegraphics[width=0.08\textwidth]{lunwen/result/img/0_gh_69.png} & 
\includegraphics[width=0.08\textwidth]{lunwen/result/label/0_gh_69.png} & 
\includegraphics[width=0.08\textwidth]{lunwen/result/heat/our/0_gh_69.png} & 
\includegraphics[width=0.08\textwidth]{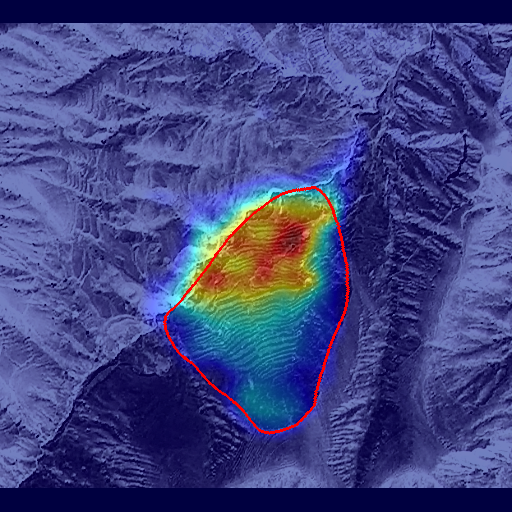} & 
\includegraphics[width=0.08\textwidth]{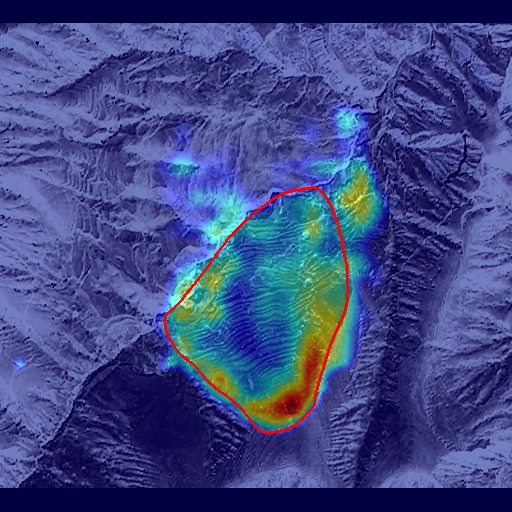} \\[-2pt]
%第2行
\includegraphics[width=0.08\textwidth]{lunwen/result/img/1_ltao_137.png} & 
\includegraphics[width=0.08\textwidth]{lunwen/result/label/1_ltao_137.png} & 
\includegraphics[width=0.08\textwidth]{lunwen/result/heat/our/1_ltao_137.png} & 
\includegraphics[width=0.08\textwidth]{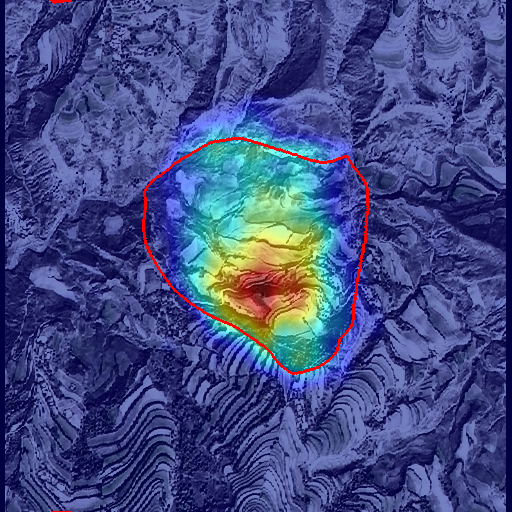} & 
\includegraphics[width=0.08\textwidth]{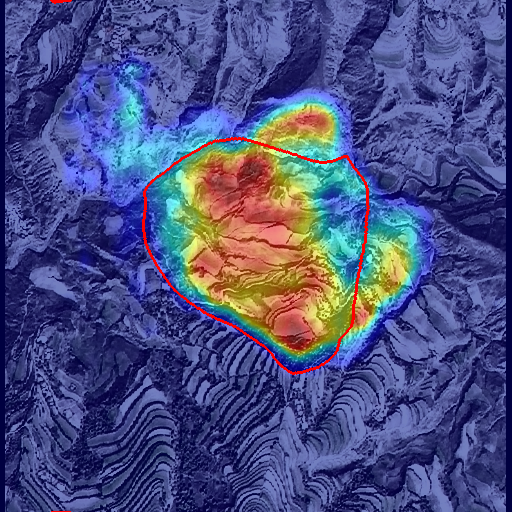} \\[-2pt]
%第3行
\includegraphics[width=0.08\textwidth]{lunwen/result/img/4_j01.png} & 
\includegraphics[width=0.08\textwidth]{lunwen/result/label/4_j01.png} & 
\includegraphics[width=0.08\textwidth]{lunwen/result/heat/our/4_j01.png} & 
\includegraphics[width=0.08\textwidth]{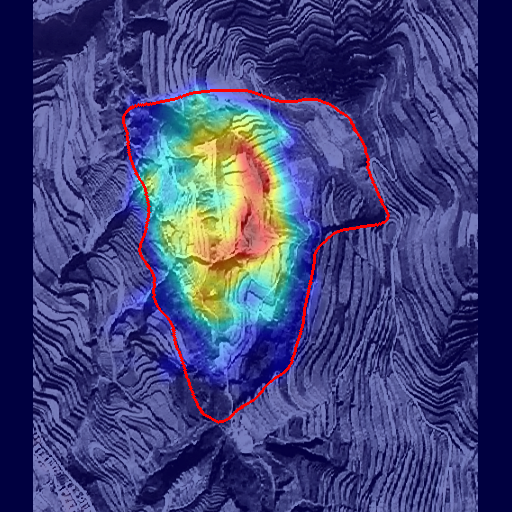} & 
\includegraphics[width=0.08\textwidth]{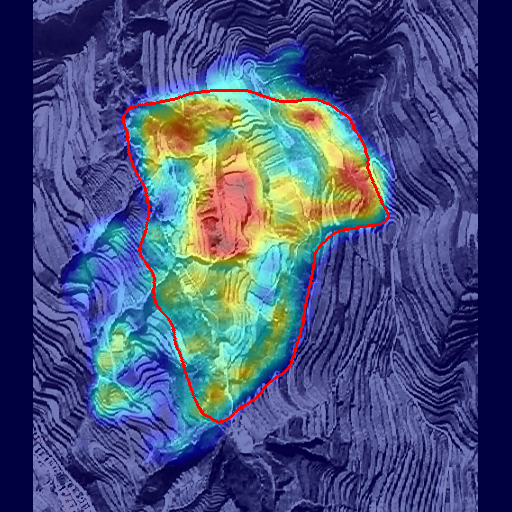} \\[-2pt]
%第4行
\includegraphics[width=0.08\textwidth]{lunwen/result/img/12_ltao_152168.png} & 
\includegraphics[width=0.08\textwidth]{lunwen/result/label/12_ltao_152168.png} & 
\includegraphics[width=0.08\textwidth]{lunwen/result/heat/our/12_ltao_152168.png} & 
\includegraphics[width=0.08\textwidth]{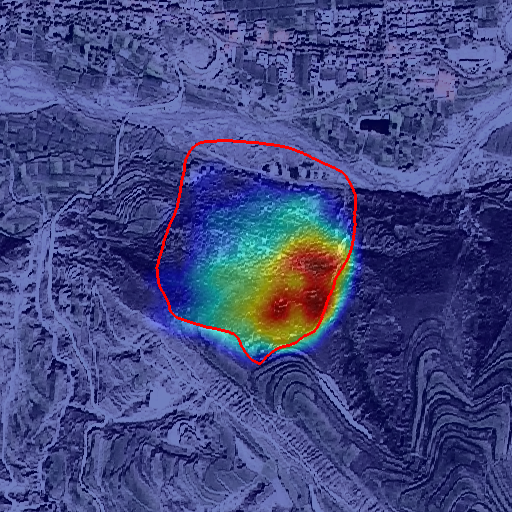} & 
\includegraphics[width=0.08\textwidth]{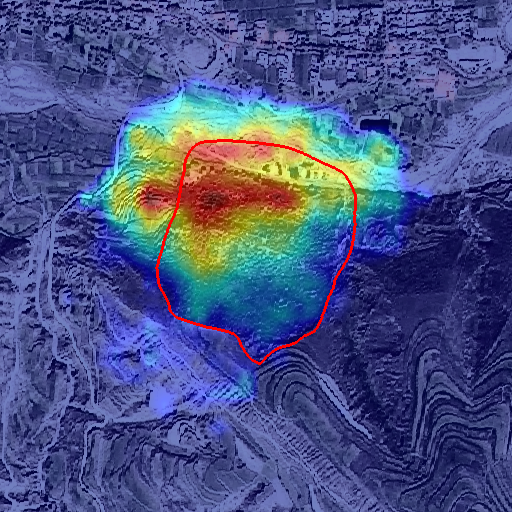} \\
(a) & (b) & (c) & (d) & (e) \\ 
\end{tabular}
\caption{Grad-CAM results. (a) Input; (b) label; (c)-(e) heat map: MRIFE, segmentation branch, and feature enhancement branch. Red line indicates the boundary.}
\label{fig_7}
\end{figure}

We conducted further analysis on the heat maps of the segmentation branch and the feature enhancement branch. As shown in Fig. \ref{fig_7}, the features extracted by the feature extraction module in segmentation branch focus more on the center of the landslide, showing differences from the features ultimately output after enhancement, particularly at the edges of the landslide. The features extracted by the feature enhancement branch are more biased towards the landslide edges. By point-to-point addition, these two sets of features are fused, supplementing the key information about the landslide, thereby improving segmentation accuracy. This analysis further validates the contribution of the feature enhancement branch to performance improvement.

\subsection{Cross-validation}
To optimize the proposed model for the best performance, we design three cross-validation experiments in this subsection: 1) The mask pixel size selection strategy; 2) The mask feature selection strategy; 3) The feature fusion method of two branches.

\begin{table}[h]
\centering
\caption{Cross-validation of the Mask Pixel Size Selection Strategy} 
\label{table_5}
\begin{tabular}{lccc}
\toprule
\multicolumn{1}{c}{Strategy} & 1-IoU & mIoU & F1-score\\
\midrule
$\mathbf{8\times8}$ & \textbf{0.3934} & \textbf{0.6680} & \textbf{0.5646} \\
$16\times16$ & 0.3648 & 0.6548 & 0.5346 \\
\bottomrule
\end{tabular}
\end{table}

\subsubsection{The Mask Pixel Size Selection Strategy}
We designed two sizes of mask pixel blocks, $8\times8$ and $16\times16$. As shown in Table \ref{table_5}, the $8\times8$ mask pixel blocks achieved the best performance. We consider that the $16\times16$ mask pixel blocks, which contain more complex information, especially at the target edge blocks, include more non-target and target-internal information, making it difficult for the masked feature modeling task to accurately extract key features corresponding to the pixel blocks. Additionally, the larger size means fewer blocks, which can also lead to difficulties in the model convergence.

\begin{figure*}[h]
\centering
\begin{tabular}{c@{\hspace{2pt}}c@{\hspace{2pt}}c@{\hspace{2pt}}c@{\hspace{2pt}}c@{\hspace{4pt}}c@{\hspace{2pt}}c@{\hspace{2pt}}c}
%第1行
\includegraphics[width=0.08\textwidth]{lunwen/result/img/0_gh_69.png} & 
\includegraphics[width=0.08\textwidth]{lunwen/result/label/0_gh_69.png} & 
\includegraphics[width=0.08\textwidth]{lunwen/result/heat/our/0_gh_69.png} & 
\includegraphics[width=0.08\textwidth]{lunwen/result/cross/B/0_gh_69.png} & 
\includegraphics[width=0.08\textwidth]{lunwen/result/cross/S/0_gh_69.png} &
\includegraphics[width=0.08\textwidth]{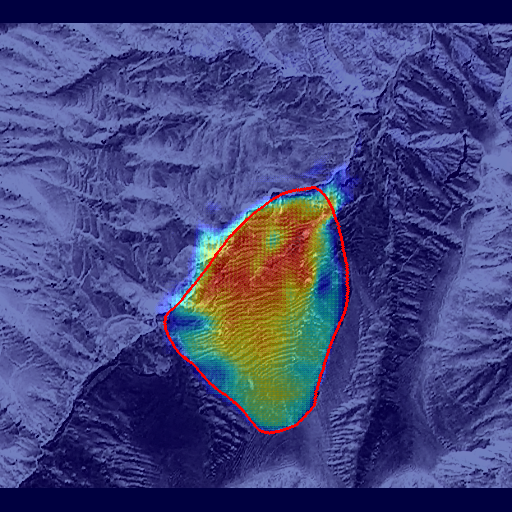} & 
\includegraphics[width=0.08\textwidth]{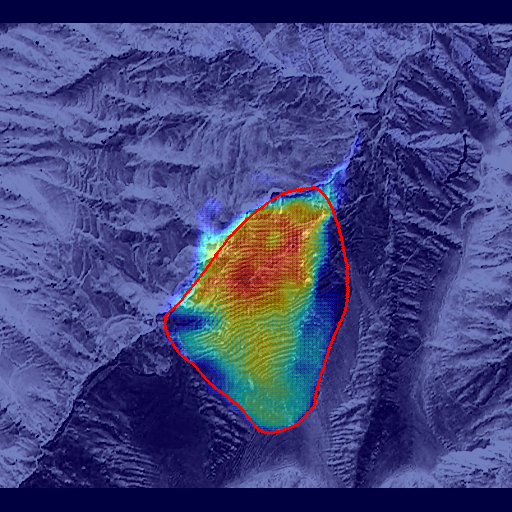} & 
\includegraphics[width=0.08\textwidth]{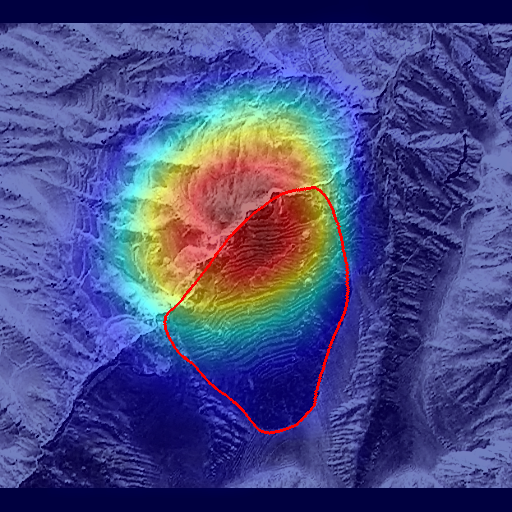} \\[-2pt]
%第2行
\includegraphics[width=0.08\textwidth]{lunwen/result/img/1_ltao_137.png} & 
\includegraphics[width=0.08\textwidth]{lunwen/result/label/1_ltao_137.png} & 
\includegraphics[width=0.08\textwidth]{lunwen/result/heat/our/1_ltao_137.png} & 
\includegraphics[width=0.08\textwidth]{lunwen/result/cross/B/1_ltao_137.png} & 
\includegraphics[width=0.08\textwidth]{lunwen/result/cross/S/1_ltao_137.png} &
\includegraphics[width=0.08\textwidth]{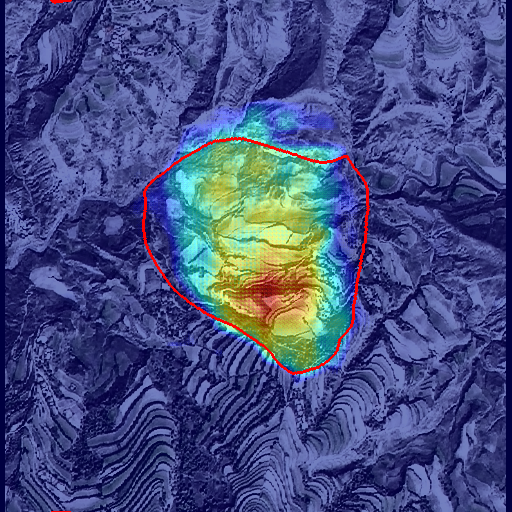} & 
\includegraphics[width=0.08\textwidth]{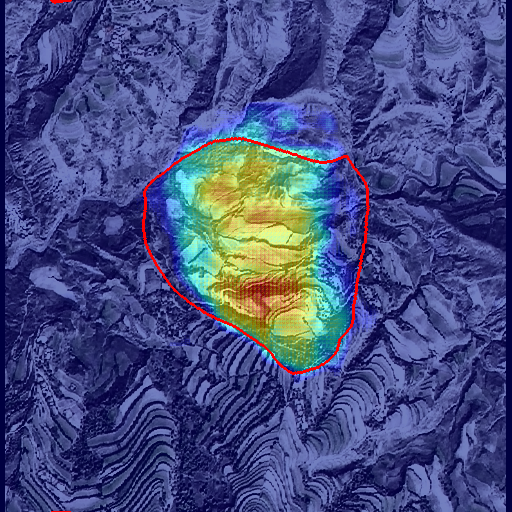} & 
\includegraphics[width=0.08\textwidth]{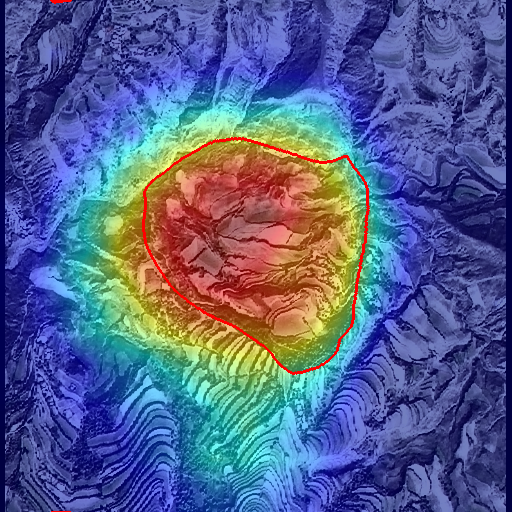} \\[-2pt]
%第3行
\includegraphics[width=0.08\textwidth]{lunwen/result/img/4_j01.png} & 
\includegraphics[width=0.08\textwidth]{lunwen/result/label/4_j01.png} & 
\includegraphics[width=0.08\textwidth]{lunwen/result/heat/our/4_j01.png} & 
\includegraphics[width=0.08\textwidth]{lunwen/result/cross/B/4_j01.png} & 
\includegraphics[width=0.08\textwidth]{lunwen/result/cross/S/4_j01.png} &
\includegraphics[width=0.08\textwidth]{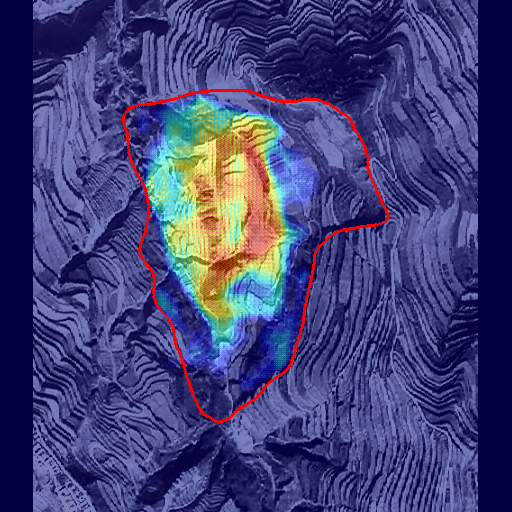} & 
\includegraphics[width=0.08\textwidth]{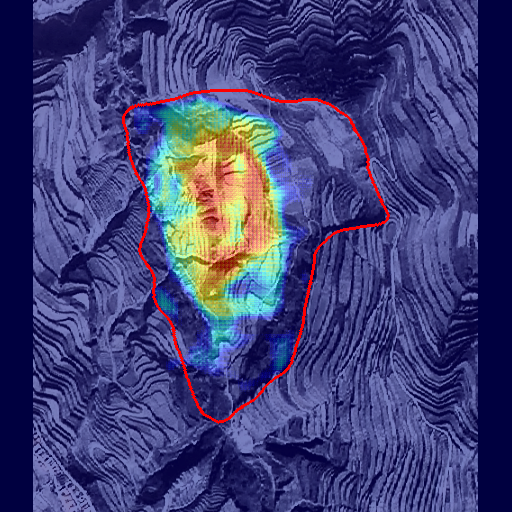} & 
\includegraphics[width=0.08\textwidth]{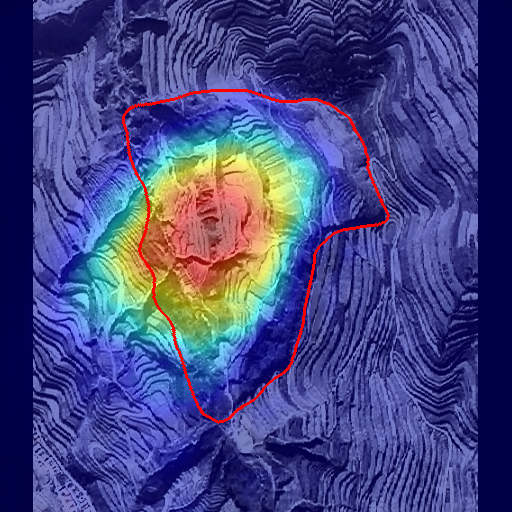} \\[-2pt]
%第4行
\includegraphics[width=0.08\textwidth]{lunwen/result/img/12_ltao_152168.png} & 
\includegraphics[width=0.08\textwidth]{lunwen/result/label/12_ltao_152168.png} & 
\includegraphics[width=0.08\textwidth]{lunwen/result/heat/our/12_ltao_152168.png} & 
\includegraphics[width=0.08\textwidth]{lunwen/result/cross/B/12_ltao_152168.png} & 
\includegraphics[width=0.08\textwidth]{lunwen/result/cross/S/12_ltao_152168.png} &
\includegraphics[width=0.08\textwidth]{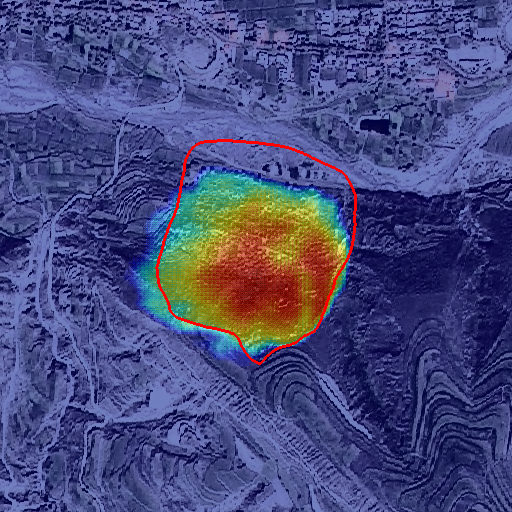} & 
\includegraphics[width=0.08\textwidth]{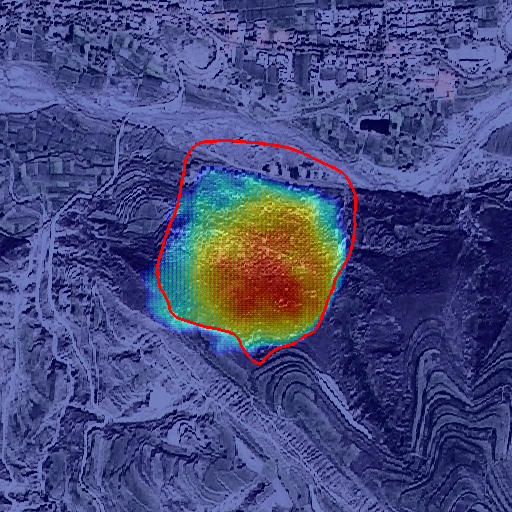} & 
\includegraphics[width=0.08\textwidth]{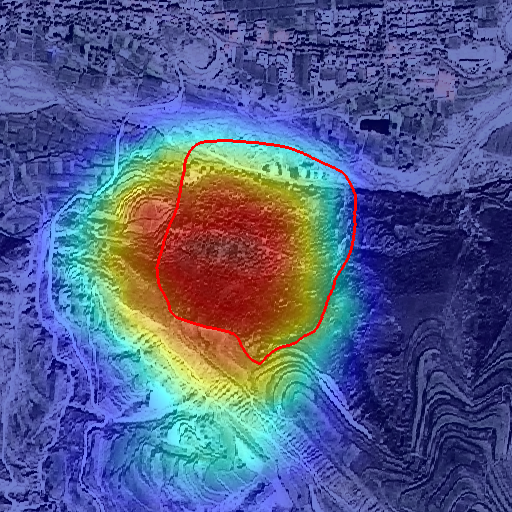} \\
(a) & (b) & (c) & (d) & (e) & (f) & (g) & (h)\\ 
\end{tabular}
\caption{Grad-CAM results. (a) Input; (b) label; (c)-(e) \textcolor{blue}{point-to-point} addition: segmentation branch after fusion, segmentation branch before fusion, and feature enhancement branch. Red line \textcolor{blue}{indicates} the boundary; (f)-(h) channel concatenation and \textcolor{blue}{weighting}: segmentation branch after fusion, segmentation branch before fusion, and feature enhancement branch. Red line indicates the boundary.}
\label{fig_8}
\end{figure*}

\subsubsection{The Mask Feature Selection Strategy}
We categorize the pixel blocks into three types: background blocks with less than $10\%$ landslide pixels, interior blocks with more than $90\%$ landslide pixels, and the remaining landslide edge blocks. These categories correspond to background features, landslide interior features, and landslide edge features, respectively. 

\begin{table}[h]
\centering
\caption{Cross-validation of the Mask Feature Selection Strategy} 
\label{table_6}
\begin{tabular}{lccc}
\toprule
\multicolumn{1}{c}{Strategy} & 1-IoU & mIoU & F1-score\\
\midrule
\textbf{Edge and Back} & \textbf{0.3934} & \textbf{0.6680} & \textbf{0.5646} \\
Center and Back & 0.3820 & 0.6621 & 0.5528 \\
All & 0.3893 & 0.6640 & 0.5604 \\
\bottomrule
\end{tabular}
\end{table}

We design three mask feature selection strategies to perform the masked feature modeling task and the semantic feature contrast enhancement task: edge and background, interior and background, and a mixture of all three. As shown in Table \ref{table_6}, the edge and background strategy achieves the best performance, while the interior and background strategy \textcolor{blue}{performs} the worst. These results confirm that the interior blocks of landslides, which exhibit features highly similar to non-landslide blocks in images, contribute minimally to landslide identification, whereas the edge features are crucial.

\subsubsection{The Feature Fusion Method of Two Branches}
To achieve dual-branch interactive feature enhancement, we designed two feature fusion methods. One method involves concatenating the feature maps of the two branches along the channel dimension, followed by compression and weighting through a $1\times1$ convolution. The other method adds the two feature maps point-to-point. The experimental heat map visualization results, as shown in Fig. \ref{fig_8}, reveal that the channel concatenation and weighting method results in obvious overfitting in the feature enhancement branch. Furthermore, comparing the feature maps of the segmentation branch before and after fusion, the feature enhancement branch almost does not bring any positive gain. In contrast, the point-to-point addition effectively utilizes the segmentation branch to constrain and guide the feature enhancement, addressing the issue of overfitting, while also enhancing the feature maps of the segmentation branch, thereby achieving interaction between the two branches.

\section{Conclusion}
This study introduces MRIFE, a novel approach for relic landslide detection, which effectively extracts key features from HRSI data and enables reliable detection of visually ambiguous relic landslides. The proposed MRIFE employs a teacher-student-based feature extraction and separation method, designed around the MFM and SFCE tasks. By leveraging supervised contrastive learning and reconstructing the edges and backgrounds of landslide targets, the model facilitates feature separation in the semantic space, enhancing both feature extraction and foreground-background discrimination. Additionally, an efficient dual-branch feature enhancement framework is developed. Through multi-task training, the two branches independently extract features while guiding and constraining each other, promoting feature enhancement and fusion. This approach not only extracts more informative features but also mitigates overfitting. Extensive experimental evaluation on real-world datasets demonstrates that MRIFE significantly improves landslide detection performance, particularly in pixel classification accuracy and edge shape precision.

\bibliographystyle{IEEEtranN}
\bibliography{reference}

\end{document}